\documentclass[10pt,twocolumn,letterpaper]{article}

\usepackage[pagenumbers]{wacv}

\usepackage[utf8]{inputenc}
\usepackage[T1]{fontenc}
\ifdefined\XeTeXversion
  \usepackage{newunicodechar}
  \newunicodechar{‘}{`}
  \newunicodechar{’}{'}
  \newunicodechar{“}{``}
  \newunicodechar{”}{''}
\fi
\usepackage{amsfonts}
\usepackage{nicefrac}
\usepackage{microtype}
\usepackage{dsfont}
\usepackage{algorithm}
\usepackage{algorithmic}
\usepackage{multirow}
\usepackage{makecell}
\usepackage{float}
\usepackage{listings}
\usepackage{tcolorbox}
\usepackage{wrapfig}
\usepackage{soul}
\usepackage{comment}
\usepackage{xurl}

\tcbuselibrary{breakable,skins}
\newcommand{\method}{TUE-Detector}
\newcommand{\toolname}[1]{\nolinkurl{#1}}

\definecolor{thinkcolor}{RGB}{52,101,164}
\definecolor{toolcolor}{RGB}{78,154,6}
\definecolor{resultcolor}{RGB}{196,160,0}
\definecolor{answercolor}{RGB}{204,0,0}
\definecolor{codebg}{RGB}{245,245,245}

\lstdefinestyle{prompt}{
  basicstyle=\ttfamily\footnotesize,
  backgroundcolor=\color{codebg},
  breaklines=true,
  breakatwhitespace=true,
  breakindent=0pt,
  columns=flexible,
  keepspaces=true,
  frame=single,
  rulecolor=\color{gray!30},
  framesep=4pt,
  framerule=0.4pt,
  showstringspaces=false,
  showspaces=false,
  showtabs=false,
  xleftmargin=0pt,
  xrightmargin=0pt,
  aboveskip=0.6em,
  belowskip=0.6em,
}

\makeatletter
\@ifundefined{nolinenumbers}{\newenvironment{nolinenumbers}{}{}}{}
\makeatother
\BeforeBeginEnvironment{lstlisting}{\begin{nolinenumbers}}
\AfterEndEnvironment{lstlisting}{\end{nolinenumbers}}

\definecolor{wacvblue}{rgb}{0.21,0.49,0.74}
\usepackage[pagebackref,breaklinks,colorlinks,allcolors=wacvblue]{hyperref}

\def\wacvPaperID{}
\def\confName{WACV}
\def\confYear{2027}

\title{TUE-Detector: A Tool-Using Expert MLLM-Based Detector for AI-Generated Videos}

\author{%
  Yichen Wu$^{1}$ \quad
  Haoxuan Qu$^{2}$ \quad
  Yongxing Dai$^{3}$ \quad
  Yan Bai$^{3}$ \\[2pt]
  Yihang Lou$^{3}$ \quad
  Yuqi Lin$^{1}$ \quad
  Hossein Rahmani$^{2}$ \quad
  Jun Liu$^{2}$ \\[2pt]
  \makebox[0pt][c]{%
    $^{1}$University of Nottingham \quad
    $^{2}$Lancaster University \quad
    $^{3}$Peking University
  }%
  \smash{\raisebox{-16pt}[0pt][0pt]{%
    \makebox[0pt][c]{\small\textbf{Code:} \href{https://github.com/Louis-YW/TUE}{github.com/Louis-YW/TUE}}%
  }}
}

\begin{document}

\maketitle
% Keep the public title block compact so the main paper remains within eight pages.
\vspace{-24pt}

\begin{abstract}
AI-generated video detection, which aims to distinguish AI-generated videos from real ones, has recently received increasing research attention. To perform this task reliably, a key challenge lies in accurately identifying subtle-yet-measurable unnatural artifacts. In this work, we address this challenge from a novel perspective of tool-mediated evidence discovery and propose \textbf{T}ool-\textbf{U}sing \textbf{E}xpert MLLM-based AI-generated Video \textbf{Detector} (\textbf{TUE-Detector}), a novel framework for AI-generated video detection. TUE-Detector trains a general MLLM into a task-tailored tool-using expert detector that learns to invoke suitable tools, collect concrete evidence of unnaturalness, and reason over the evidence for reliable detection. Meanwhile, TUE-Detector further introduces novel designs to equip the expert detector with high-quality and suitable tools. Extensive experiments demonstrate the effectiveness of our framework.
\end{abstract}

%=============================================================================
\section{Introduction}
\label{sec:intro}
%=============================================================================

Recent AI video generation models can produce increasingly realistic videos. This ability has benefited many applications, such as content creation, education, and entertainment. However, it also brings new risks. Since AI-generated videos can now look very similar to real videos, malicious users may use them to imitate real people or fabricate events, causing harms such as political misinformation~\cite{dobber2021}, non-consensual intimate content~\cite{lazard2025}, and commercial fraud~\cite{muhly2025}. Therefore, it is important to determine whether a given video is real or AI-generated, and AI-generated video detection has thus received substantial research attention recently~\cite{vidguard2025,videoversitas2026,skyra2026,genbuster2025,restrav2024,mare2026}.  In particular, motivated by the rich visual knowledge and strong reasoning ability of multimodal large language models (MLLMs), recent studies~\cite{skyra2026,genbuster2025,vidguard2025,videoversitas2026} have increasingly explored MLLM-based approaches as a promising direction for this task.

\begin{figure}[t]
\centering
\includegraphics[width=\linewidth]{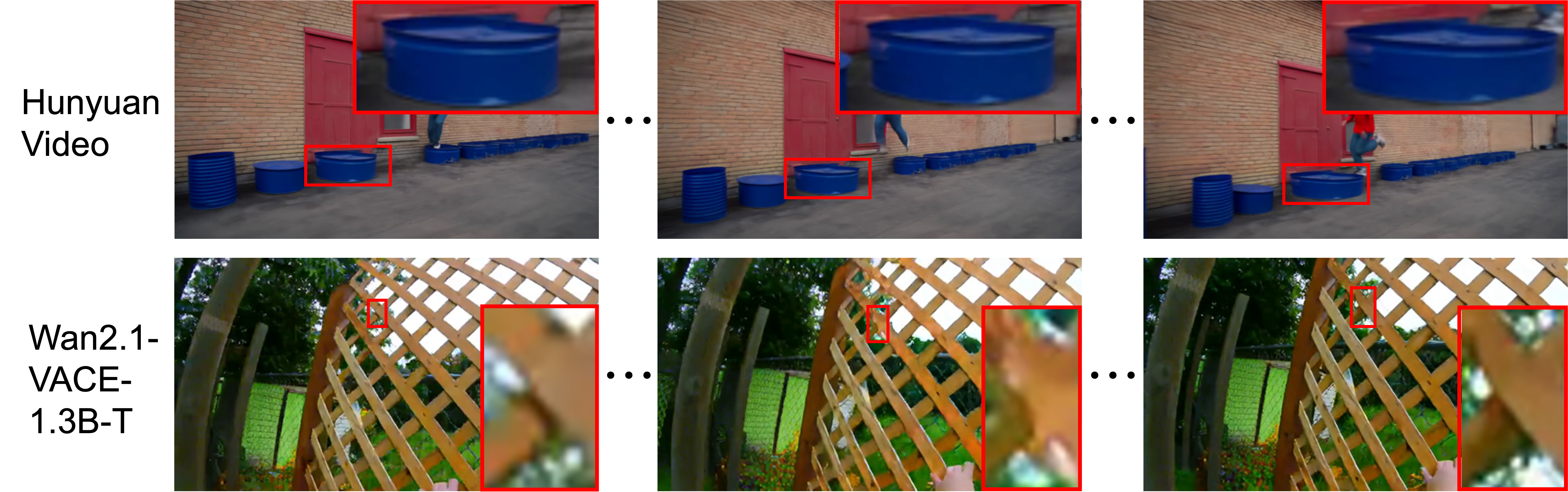}
\caption{Illustration of videos generated by recent AI video generation models. Recent AI-generated videos can already appear highly similar to real videos, with remaining differences mainly lying in subtle-yet-measurable unnatural artifacts, such as subtle shape changes in the upper example and slight texture alterations in the lower example, highlighted by \textcolor{red}{red} boxes.}
\label{fig:tool_examples}
\end{figure}

% Illustration of videos generated by recent AI video generated models. As shown, 他们已经looks real, whose differences和真实的视频已经主要坐落于subtle-yetmeasurable unnatural artifects (shape change in the 上面 and texture alter in the bottom)

To perform AI-generated video detection reliably, recent works~\cite{davidxr12025,lin2025brokenvideos,genvideo2024} have increasingly highlighted the importance of identifying \textit{subtle-yet-measurable unnatural artifacts}: unobtrusive cues that are difficult to notice at first glance but provide useful evidence for distinguishing real from AI-generated videos. These artifacts often appear as subtle unnaturalness in measurable visual or temporal properties, such as shape, texture, motion, position, color, and boundary stability. For example, AI-generated videos may contain subtle shape changes, slight texture alterations, mildly unnatural object motion, or local lighting and shadow inconsistencies. As illustrated in Figure~\ref{fig:tool_examples}, as video generation models become stronger, obvious artifacts become less frequent, and the differences between real and AI-generated videos increasingly appear as such subtle-yet-measurable cues~\cite{ni2026genvidbench,waverep2024}. Meanwhile, existing detection methods have become relatively effective at capturing salient artifacts when they are present~\cite{aigvdet2024,vahdati2024beyond}. Therefore, accurately identifying subtle-yet-measurable unnatural artifacts has become an increasingly important problem for further improving AI-generated video detection~\cite{decof2024,he2024exposing,genvideo2024,restrav2024}. However, this remains challenging because these artifacts are not only subtle, but also highly diverse across video categories, scene contents, and generation models.

\begingroup\looseness=-1
To handle this challenge, we first note that identifying subtle-yet-measurable unnatural artifacts can be naturally viewed as an evidence discovery problem. Specifically, the detector needs to analyze diverse subtle cues in a video and examine whether they constitute concrete evidence of unnaturalness. Once such evidence is identified and verified, the detector can further reason over it to determine whether the video is real or AI-generated. This perspective motivates us to consider how humans handle similar evidence-based visual analysis problems. In particular, we note that, when facing subtle visual cues, trained human analysts usually do not rely solely on direct visual impressions. Instead, through training, they learn when and how to use suitable tools to examine different types of subtle visual cues~\cite{enfsi2018five,jerian2007forensic,fontani2013forensictool}. For example, they may use contour-inspection tools to measure subtle shape changes, texture-analysis tools to examine slight texture alterations, and lighting-probe tools to assess local lighting and shadows. After collecting such concrete evidence from different regions and frames of a video, they can reason over the evidence and make a final judgment.
\par\endgroup

\begingroup\looseness=-1
Motivated by this observation, to better identify and analyze subtle-yet-measurable unnatural artifacts in videos for more reliable AI-generated video detection, we aim to approach this task from the perspective of tool-mediated evidence discovery. Specifically, we aim to equip MLLMs with suitable tools and transform them into tool-using experts, enabling them to collect concrete evidence of unnaturalness and reason over such evidence for reliable detection. To achieve this goal, given a general MLLM, one potential solution is to provide it with a heuristically designed toolset and directly prompt it to use these tools for AI-generated video detection. Recently, several works~\cite{factguard2026,forgeryvcr2026,lavid2025,fakehunter2025} have explored similar ideas by instructing general MLLMs to use external tools for various visual detection and verification tasks. However, a general MLLM is not naturally a tool-using expert. As suggested by prior studies~\cite{toolllm2023,llavaplus2024}, directly prompting a general MLLM to use tools does not necessarily mean that it understands when a tool should be invoked, under what conditions it is useful, or how its output should be incorporated into task-specific reasoning. This is undesirable for our task, where accurate AI-generated video detection calls for a detector model that can adaptively select and properly use suitable tools to discover diverse subtle-yet-measurable unnatural artifacts.
\par\endgroup

\begingroup\looseness=-1
To this end, instead of directly combining a general MLLM with a heuristic toolset, we aim to train the general MLLM into a task-tailored tool-using expert, while further equipping it with suitable tools for AI-generated video detection. Achieving this goal, however, is non-trivial due to two key challenges. First, obtaining suitable tools for this task is difficult. AI-generated videos may exhibit diverse subtle artifacts across different video categories, scenes, and generation models, making it hard to determine in advance which tools are truly useful and suitable for this task. Second, even with a candidate toolset, teaching MLLMs to use tools effectively remains challenging. Different videos may contain different subtle cues in different regions, frames, and temporal scales, so effective tool use cannot follow a fixed pattern. Instead, the model needs to adaptively decide which tools to use, where and when to apply them, and how to incorporate their outputs into detection reasoning, so that the collected evidence can support reliable detection. To address the above challenges, we propose \textbf{T}ool-\textbf{U}sing \textbf{E}xpert MLLM-based AI-generated Video \textbf{Detector} (\textbf{TUE-Detector}), a novel framework for evidence-grounded AI-generated video detection. To the best of our knowledge, TUE-Detector is the first framework that trains a general MLLM into a task-tailored tool-using expert detector for AI-generated video detection, while further equipping it with suitable and useful tools for identifying subtle-yet-measurable unnatural artifacts. Below, we outline our framework.
\par\endgroup

\begingroup\looseness=-1
Overall, TUE-Detector performs AI-generated video detection through tool-mediated evidence discovery: it learns to invoke diverse tools to examine subtle cues, collect concrete evidence of unnaturalness, and reason over the evidence for the final judgment. However, when starting from a general MLLM, neither a suitable toolset nor the corresponding tool-use ability is readily available. Directly optimizing both from scratch can thus be difficult. To tackle this, we design a progressive two-stage training process that first builds basic tool-use ability with a fixed heuristic toolset, and then further improves both the detector and the toolset. Specifically, in the first stage, we construct an initial heuristic toolset and keep it fixed. With the toolset fixed, we focus on helping the MLLM acquire basic tool-use ability for AI-generated video detection. In particular, to make this stage effective, we propose an \textbf{influence-based teacher-knowledge guidance strategy}, which provides high-quality guidance for learning tool use and evidence-based reasoning. However, since the toolset in this stage is heuristically constructed and remains fixed during training, both the model's tool-use ability and the toolset itself still have room for further improvement. Thus, after the first stage, we introduce a second stage, in which we further optimize both the detector and the toolset. In particular, in this stage, to effectively improve the usability and suitability of the toolset, we propose a \textbf{tool-use experience-guided tool evolution strategy}. Through this progressive process, TUE-Detector is ultimately equipped with suitable tools and strong tool-use ability, driving its effective identification of subtle-yet-measurable unnatural artifacts for reliable AI-generated video detection.
\par\endgroup

\begingroup\looseness=-1
Our contributions are: 1) We propose TUE-Detector, a novel framework for AI-generated video detection. To the best of our knowledge, this is the first work to explore training a general MLLM into a task-tailored tool-using expert detector for AI-generated video detection. 2) We introduce several designs in TUE-Detector to equip it with high-quality and suitable tools, as well as strong tool-use capability. 3) Our method achieves superior performance on the evaluated benchmarks.
\par\endgroup

%=============================================================================
\section{Related Work}
\label{sec:related}
%=============================================================================

\noindent\textbf{AI-generated video detection.} Owing to the rapid advancement of AI video generation models~\cite{sora2024,runway2024,kling2024}, AI-generated video detection has received extensive research attention~\cite{aigvdet2024,decof2024,genvideo2024,d3detector2024,restrav2024,nsgvd2025,genconvit2023,styleflow2024,fakestormer2025,sta4deepfake2025,unite2025,commonsense2024,mare2026,videoversitas2026,vidguard2025,edvdllama2025,exddv2025,skyra2026,genbuster2025,busterx2025,ivyfake2025,davidxr12025}. 
Early studies explored various neural architectures for this task, such as CNN-based methods~\cite{aigvdet2024,decof2024,d3detector2024,restrav2024} and RNN-based methods~\cite{rcn2019,ftcn2021,styleflow2024}. 
With the advent of MLLMs, recent works have increasingly leveraged their visual understanding and reasoning capability for AI-generated video detection~\cite{commonsense2024,unite2025,skyra2026,genbuster2025,busterx2025,ivyfake2025,davidxr12025,vidguard2025}. 
For example, Skyra~\cite{skyra2026} trains a specialized MLLM to identify human-perceivable visual artifacts as grounded evidence for both detection and explanation. 
VidGuard-R1~\cite{vidguard2025} introduces a GRPO-based reinforcement learning framework with specialized reward models, encouraging the model to discover temporal and physics-grounded artifacts for more reliable authenticity reasoning. 
BusterX~\cite{busterx2025} formulates AI-generated video detection as a visual reasoning task and develops an RL-trained MLLM detector.

Different from existing studies, to the best of our knowledge, this work is the first to explore training a general MLLM into a task-tailored tool-using expert detector for AI-generated video detection.

\noindent\textbf{Tool usage.}
\begingroup\looseness=-1
Recently, the idea of using external tools to assist task solving has been studied in various tasks~\cite{toolformer2023,webgpt2021,pal2023,react2023,art2023,lavid2025,fakehunter2025,multimediaverif2025,aifo2025,forenagent2025,evoguard2026}, such as open-domain question answering~\cite{webgpt2021}, program-aided mathematical reasoning~\cite{pal2023}, fact verification~\cite{react2023}, and image forensic analysis~\cite{aifo2025,forenagent2025,evoguard2026}.
Different from these existing works, we design a novel framework that trains a general MLLM into a task-tailored tool-using expert detector, enabling it to effectively identify subtle-yet-measurable unnatural artifacts and handle the AI-generated video detection task.
\par\endgroup

%=============================================================================
\section{Method}
\label{sec:method}
%=============================================================================

\begingroup\looseness=-1
To perform reliable AI-generated video detection, a key challenge lies in accurately identifying subtle-yet-measurable unnatural artifacts. To better tackle this challenge, we propose TUE-Detector, a novel framework that trains a general MLLM into a task-tailored tool-using expert detector. Instead of a learned aggregation of outputs from MLLM and tools (as shown in Appendix~\ref{app:output_aggregation}), TUE-Detector learns to invoke suitable tools, collect concrete evidence, and reason over it to determine whether a video is real or AI-generated. To achieve this, TUE-Detector follows a two-stage training process. In Stage 1 (Sec.~\ref{sec:stage1}), we construct an initial heuristic toolset and keep it fixed, allowing training to focus on equipping the general MLLM with basic tool-use and evidence-based reasoning ability. Stage 2 (Sec.~\ref{sec:stage2}) then starts from the preliminary detector and heuristic toolset obtained in Stage 1, and further improves both, leading to stronger tool-use ability and a more task-suitable toolset.
\par\endgroup

\begin{figure}[t]
\centering
\includegraphics[width=\columnwidth]{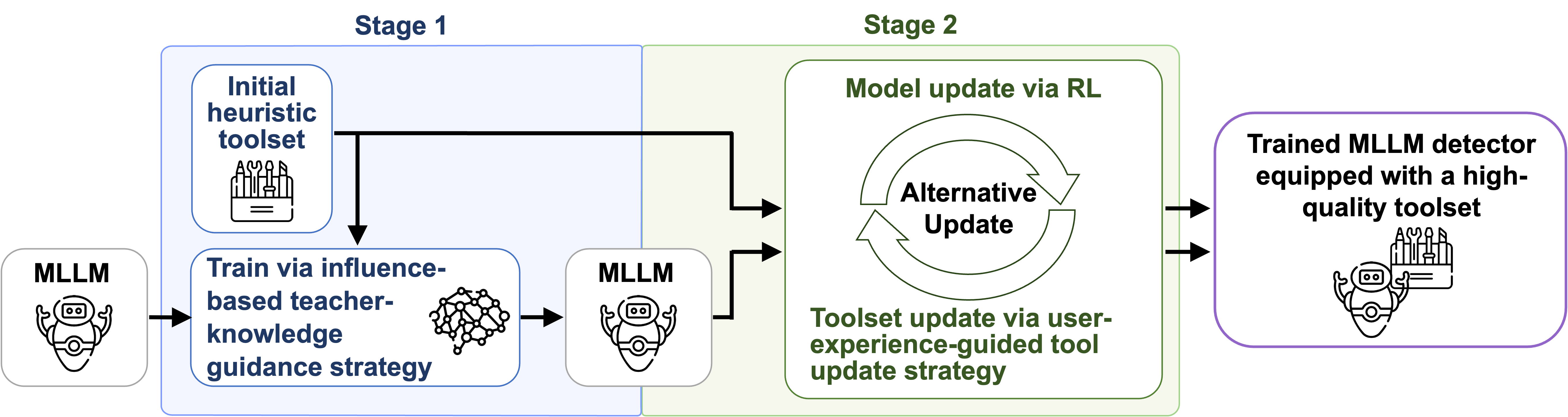}
\caption{Illustration of the overall training process of our framework.}
\label{fig:overview}
\end{figure}

\subsection{Training Stage 1}
\label{sec:stage1}
%-----------------------------------------------------------------------------

\begingroup\looseness=-1
In this stage, we aim to construct an initial heuristic toolset and train the general MLLM to acquire basic tool-use ability under this fixed toolset. Specifically, after constructing the initial toolset for AI-generated video detection, we train the model to use these tools for tool-mediated evidence discovery and evidence-based reasoning. To make the model effectively acquire this ability, we propose an \textbf{influence-based teacher-knowledge guidance strategy}. Below, we first describe how we construct the heuristic toolset, and then introduce this strategy.
\par\endgroup

\noindent\textbf{Initial Heuristic Toolset.} Our framework first uses an MLLM-based automated process to heuristically construct a fixed initial toolset for tool-mediated evidence discovery. In this process, since AI-generated videos may contain diverse subtle-yet-measurable unnatural artifacts, the construction is guided to include tools that may help collect different types of evidence potentially useful for AI-generated video detection. Following this principle, an initial toolset containing thirteen tools is built. Each tool exposes a common Python-callable interface, while its implementation may orchestrate pretrained segmentation, tracking, OCR, spectral/frequency, temporal, or frame-level analysis backends and serialize their outputs as structured evidence. These tools can support the examination of possible evidence such as illumination inconsistency, implausible object states, and abnormal textual regions. More details of both the tool construction process and the resulting toolset are provided in Appendix~\ref{app:toolset}.

\noindent\textbf{Influence-based Teacher-knowledge Guidance Strategy.} Given the fixed initial toolset, the goal of this strategy is to help the general MLLM effectively acquire basic tool-use ability for AI-generated video detection. To achieve this, inspired by the observation that high-quality references can provide useful guidance when a model lacks task-specific procedural knowledge~\cite{wei2022finetuned,toolllm2023,llavaplus2024}, we aim to provide the model with reference trajectories for tool-mediated evidence discovery and evidence-based reasoning. Ideally, such references should specify which tools to invoke, what concrete evidence of unnaturalness can be obtained from the tool outputs, and how the collected evidence should be reasoned over to reach the final detection verdict. In this way, the model can learn not only the final verdict, but also the intermediate behaviours required for tool-mediated detection, including tool selection, evidence collection, tool-output interpretation, and evidence-grounded reasoning. However, obtaining such reference trajectories can be difficult in practice, because it requires detailed process-level annotations of tool use, evidence collection, tool-output interpretation, and reasoning. In many existing AI-generated video detection datasets, each training video is usually provided with only a binary real/fake label, rather than such detailed references. This creates a fundamental supervision gap: \textit{the desired reference trajectories are useful for learning tool-mediated evidence discovery, but they are usually unavailable in training data}.

\noindent\ul{Initial reference construction.} To address the above supervision gap and obtain candidate reference trajectories, we draw inspiration from the capability of existing API-accessible MLLMs. Although these models are not tailored for AI-generated video detection, they often possess broader general knowledge and stronger instruction-following ability than the open-source MLLM to be trained. While they may not produce reliable trajectories every time, their stronger capabilities and stochastic generation make it possible to obtain useful tool-use and reasoning behaviours from multiple attempts. Therefore, we use API-accessible MLLMs as noisy teachers to provide candidate references for tool-mediated evidence discovery and evidence-based reasoning. Specifically, based on this insight, we adopt a generate-then-filter procedure. For each training video, we prompt a capable API-accessible MLLM with sampled video frames, the fixed initial toolset, and the detection instruction. The teacher is required to generate complete tool-augmented reasoning trajectories, including tool selection, tool calls, tool observations, evidence interpretation, and the final detection verdict (more details in Appendix~\ref{app:teacher_cot}). Since these trajectories are generated by a noisy teacher, we use the available real/fake label as an outcome-level filtering signal and remove trajectories whose final verdict is inconsistent with the label. This gives an initial reference dataset $\mathcal{S}_{\text{ref}} = \{(v_j, \tau_j^*)\}$, where $v_j$ is a training video and $\tau_j^*$ is a retained teacher-generated trajectory whose final verdict matches the label.

To guide the model to imitate the retained tool-use and reasoning behaviours, we then aim to train it based on these trajectories. However, each trajectory contains both active tokens generated by the teacher and observations returned by external tools. Because of this, directly optimising all tokens would encourage the model to memorise tool observations and weaken its learning of the actual tool-use behaviours, such as reasoning steps and tool-call generation. Therefore, inspired by Tool-Star~\cite{toolstar2025}, for each training sample, we define the masked token-wise cross-entropy loss as:
\begin{equation}
\setlength{\abovedisplayskip}{3pt}
\setlength{\belowdisplayskip}{3pt}
\mathcal{L}_j(\vartheta)
=
-\sum_{\eta}
\log p_\vartheta(\sigma_{\eta} \mid v_j, \widehat{\mathcal{I}}, \sigma_{<\eta}) \cdot m_{\eta},
\label{eq:naive_sft}
\end{equation}
where $\sigma_{\eta}$ is the $\eta$-th token in trajectory $\tau_j^*$, $\vartheta$ denotes the detector parameters, and $p_\vartheta(\sigma_{\eta} \mid v_j, \widehat{\mathcal{I}}, \sigma_{<\eta})$ is the autoregressive probability of generating token $\sigma_{\eta}$ given the video $v_j$, instruction $\widehat{\mathcal{I}}$, and previous tokens $\sigma_{<\eta}$. Here, $m_{\eta} \in \{0,1\}$ is a token-level mask, set to $1$ for tokens corresponding to active reasoning and tool-call generation, and set to $0$ for tokens corresponding to tool results. This focuses learning on tool-use decisions and evidence-based reasoning, rather than copying tool observations.

\noindent\ul{Remaining challenge.} As described above, we construct initial reference trajectories through a generate-then-filter procedure, where the binary real/fake label is used to remove trajectories with incorrect final verdicts. Yet, this does not guarantee that the retained trajectories are all high-quality references, because the filtering process only checks the final outcome rather than the intermediate tool-use process. This means that a retained trajectory may match the label while relying on incorrect reasoning, invoking unsuitable tools, missing important evidence, or misinterpreting tool results. Thus, to better learn from these imperfect teacher-generated trajectories, we need to determine which retained trajectories are truly useful for training, rather than treating all of them equally.

\noindent\ul{Influence-based trajectory weighting.}
To handle the above challenge, we need a measure that tells whether learning from a retained trajectory is beneficial for the detector. Influence functions~\cite{koh2017understanding,datainf2023,bhatt2021fast,ren2020not} provide such a measure by estimating how the validation loss would change if a training sample were up-weighted. This matches our goal: a useful teacher-generated trajectory should guide the model towards better performance on held-out validation data. Specifically, given a held-out validation set, the influence score $\chi_j$ of each retained trajectory $\tau_j^*$ can be defined based on the standard influence-function formulation as:
\begin{equation}
\setlength{\abovedisplayskip}{3pt}
\setlength{\belowdisplayskip}{3pt}
\chi_j
=
-\nabla_\vartheta \mathcal{L}_{\text{val}}(\vartheta)^\top
\mathcal{H}_\vartheta^{-1}
\nabla_\vartheta \mathcal{L}_j(\vartheta),
\label{eq:inf}
\end{equation}
where $\mathcal{L}_{\text{val}}(\vartheta)$ is the averaged loss on the held-out validation set, $\mathcal{L}_j(\vartheta)$ is the masked training loss for imitating trajectory $\tau_j^*$ on video $v_j$, and $\mathcal{H}_\vartheta$ is the Hessian of the averaged training loss.
In our context, this score reflects whether imitating a retained teacher-generated trajectory would improve or hurt validation performance, and thus serves as a proxy for trajectory utility.

However, directly applying Eq.~\ref{eq:inf} to MLLM detector training in our context appears difficult for two reasons. First, it requires the inverse Hessian $\mathcal{H}_\vartheta^{-1}$ over the parameters of a large MLLM. Since $\mathcal{H}_\vartheta$ is a $P_\vartheta\times P_\vartheta$ matrix, where $P_\vartheta$ denotes the number of trainable model parameters, explicitly constructing or inverting it is prohibitively expensive. Second, our goal is not to evaluate one trajectory, but to score all retained teacher-generated trajectories in the reference dataset $\mathcal{S}_{\text{ref}}$. This would require many sample-specific inverse-Hessian-vector computations, making direct influence estimation difficult to scale. Together, these two issues make it challenging to apply Eq.~\ref{eq:inf} directly in our setting.

Building on the ridge-regression formulation of RRInf~\cite{rrinf2025}, we adapt influence estimation to our masked SFT objective for teacher-generated tool-use trajectories. Specifically, as detailed in Appendix~\ref{app:influence}, the influence scores of a batch of $J_b$ retained trajectories can be estimated jointly by solving the following ridge-regression problem:
% \begin{equation}
% \setlength{\abovedisplayskip}{3pt}
% \setlength{\belowdisplayskip}{3pt}
% \mbox{\qhx{$\displaystyle \widehat{\chi}_j = \hat{\xi}_j$}},
% \quad
% \hat{\xi}
% =
% \operatorname*{arg\,min}_{\xi \in \mathbb{R}^{J_b}}
% \frac{1}{P_\vartheta}\|\Psi \xi - b\|_2^2
% +
% \rho\|\xi\|_2^2,
% \label{eq:ridge}
% \end{equation}
\begin{equation}
\setlength{\abovedisplayskip}{3pt}
\setlength{\belowdisplayskip}{3pt}
\widehat{\chi}
=
\operatorname*{arg\,min}_{\xi \in \mathbb{R}^{J_b}}
\frac{1}{P_\vartheta}\|\Psi \xi - b\|_2^2
+
\rho\|\xi\|_2^2,
\label{eq:ridge}
\end{equation}
% (\qhx{Option: write $\hat{\xi}
% =
% \operatorname*{arg\,min}_{\xi \in \mathbb{R}^{J_b}}
% \frac{1}{P_\vartheta}\|\Psi \xi - b\|_2^2
% +
% \rho\|\xi\|_2^2$ and define $\widehat{\chi}:=\hat{\xi}$ in the following sentence.})
where $\widehat{\chi} \in \mathbb{R}^{J_b}$ and the $j$-th entry of $\widehat{\chi}$ serves as the estimated influence score of the $j$-th retained trajectory in the batch. In Eq.~\ref{eq:ridge}, $\Psi \in \mathbb{R}^{P_\vartheta \times J_b}$ is the matrix of per-trajectory training gradients in the batch, with each column corresponding to one retained trajectory, $b \in \mathbb{R}^{P_\vartheta}$ is the aggregated validation target, $P_\vartheta$ is the gradient dimension, and $\rho>0$ is the ridge penalty. 
More details on the construction of Eq.~\ref{eq:ridge} and its solution are provided in Appendix~\ref{app:influence}.

Reducing the influence-score measurement in Eq.~\ref{eq:inf} to the ridge-regression problem solving in Eq.~\ref{eq:ridge} offers two practical advantages. First, it removes the need to explicitly construct or invert the Hessian, addressing the main computational bottleneck in Eq.~\ref{eq:inf}. Second, the ridge-regression formulation enables batch-wise estimation of per-trajectory influence scores, avoiding separate inverse-Hessian-vector computations for individual trajectories. This substantially improves efficiency while preserving a utility score for each retained trajectory. As a result, by leveraging Eq.~\ref{eq:ridge}, influence-score estimation becomes practical in our scenario.

% Once the influence scores of the retained training trajectories are estimated, we use them to construct adaptive trajectory weights, so that trajectories with larger influence magnitudes provide stronger supervision. 
Once the influence scores of the retained training trajectories are estimated, we use them to construct adaptive trajectory weights, assigning stronger supervision to more influential trajectories.
% Specifically, we normalize the estimated influence scores into per-trajectory weights $\{w_j\}$ and replace Eq.~\ref{eq:naive_sft} with the following weighted masked token-wise cross-entropy loss for each retained trajectory:
Specifically, we normalize the estimated influence scores into per-trajectory weights ${w_j}$ and use the following weighted version of Eq.~\ref{eq:naive_sft} as the loss function for each retained trajectory:
\begin{equation}
\setlength{\abovedisplayskip}{3pt}
\setlength{\belowdisplayskip}{3pt}
\mathcal{L}_j^{w}(\vartheta)
=
-w_j
\sum_{\eta}
\log p_\vartheta(\sigma_{\eta} \mid v_j, \widehat{\mathcal{I}}, \sigma_{<\eta}) \cdot m_{\eta}.
\label{eq:sft_loss}
\end{equation}
% Under this loss, more positively influential teacher-generated trajectories contribute more to optimization, while less positively influential ones are down-weighted. 
Under this loss, teacher-generated trajectories with greater influence contribute more to optimization, while less influential ones are down-weighted.
Notably, since the influence of a trajectory may change as the model improves, we do not keep these weights fixed throughout training. Instead, we periodically recompute the estimated influence scores under the current model, update the trajectory weights, and continue training with the updated weighted loss every $\gamma$ epochs. 
This iterative reweighting adapts teacher guidance as the model gradually acquires tool-use ability.

Overall, the procedure treats the API-accessible MLLM as a noisy proposal source rather than an oracle. The binary label is used only after a trajectory has been generated to filter inconsistent trajectories; it is not included in the teacher prompt and is never available at inference. Across alternative teacher choices, the final results remain stable. These results are reported in the \hyperref[app:teacher_model]{teacher-model} and \hyperref[app:teacher_gt]{GT-label} ablations in Appendix~\ref{app:moreablations}.

% \qhx{Till here.}
% \qhx{Somewhere write heldout 10\%}

\subsection{Training Stage 2}
\label{sec:stage2}

Above, we have obtained two preliminary components: an initial heuristic toolset and a tool-using MLLM model trained through the influence-based teacher-knowledge guidance strategy. At this point, a straightforward way is to directly deploy this preliminary system for AI-generated video detection. However, both components are still limited. On the model side, although the first-stage training equips the MLLM with basic tool-use ability, the model may still overfit to demonstrated trajectory patterns rather than learning a generalizable tool-use policy. On the tool side, the initial toolset is heuristically constructed and remains static; thus, it may not be sufficiently adapted to the tool-use behaviors of the MLLM or to the diverse subtle-yet-measurable unnatural artifacts appearing in AI-generated videos. These limitations motivate a second-stage optimization process that further improves both the model and the toolset.

However, jointly optimizing them is non-trivial because the two components are heterogeneous in form. The model is parameterized as a neural network and can be optimized through gradient-based training, whereas the tools are implemented as callable Python functions for specialized visual or temporal analysis and cannot be updated in the same way. At the same time, the model and the toolset are tightly coupled: the model decides how tools are used, while the toolset determines what evidence the model can obtain. Therefore, optimizing either component in isolation is insufficient, while directly optimizing them together is difficult. To make the problem tractable, we adopt an alternating optimization strategy: we optimize the model given the current toolset, and then evolve the toolset based on the tool-use experience collected from the updated model. These two steps are performed iteratively, allowing the model and the toolset to progressively adapt to each other.

\noindent\textbf{Model Optimization.}
For model optimization, as discussed above, our main goal in this stage is to improve the generalization ability of the first-stage tool-using MLLM. Although the first-stage training equips the model with basic tool-use behaviors from teacher-generated demonstrations, it may not sufficiently encourage the model to make better tool-use decisions beyond the demonstrated patterns. Since reinforcement learning can improve models through task-level feedback and help them go beyond supervised trajectories, we thus adopt RL to optimize the MLLM in this stage.

To apply RL to our setting, we observe that in many RL optimization processes, many components are task-agnostic and can be directly applied to our scenario, such as the mechanism for deriving update directions from reward signals. Hence, adapting these RL processes to our scenario mainly requires specifying the task-dependent component, i.e., a reward function aligned with the objective of evidence-grounded AI-generated video detection. Based on this, we design a suitable reward function for our scenario below. Specifically, in our task, to enable the model to perform accurate AI-generated video detection through effective tool use, we find it useful to reward both the final detection result and the intermediate tool-use trajectory. For the result reward $\mathcal{R}_{\text{result}}$, we use a binary reward indicating whether the final result is correct, where $\mathcal{R}_{\text{result}}=1$ if the result is correct and $\mathcal{R}_{\text{result}}=0$ otherwise. For the trajectory reward $\mathcal{R}_{\text{trajectory}}$, we evaluate the tool-use trajectory quality from multiple aspects (details in Appendix~\ref{app:stage2_rl}). Collectively, this gives the following reward:
\begin{equation}
\setlength{\abovedisplayskip}{3pt}
\setlength{\belowdisplayskip}{3pt}
\mathcal{R} = \mathcal{R}_{\text{result}} + \mathcal{R}_{\text{trajectory}}.
\label{eq:reward}
\end{equation}
We provide the full model optimization details for this stage in Appendix~\ref{app:stage2_rl}. As shown in the \hyperref[app:ablation_alpha]{reward-weight} and \hyperref[app:ablation_reward_component]{reward-component} ablations in Appendix~\ref{app:moreablations}, the performance remains stable across reward-weight settings and remains strong, although lower, when either reward component is removed.

\noindent\textbf{Tool Optimization.} Having described how the model is optimized, we next introduce how to evolve the toolset. Notably, compared with model optimization, toolset optimization is more challenging. On the one hand, improving a toolset typically requires measuring the quality of each tool, which is not straightforward for custom tools used in MLLM-based reasoning. Moreover, even if a tool can be evaluated, it remains unclear how to update the tool according to this evaluation. Our tools use a common Python-callable interface, but individual implementations may wrap non-differentiable visual or temporal backends; the interface is therefore an orchestration boundary rather than a restriction to simple scalar heuristics. To address these challenges, we propose a \textbf{tool-use experience-guided tool evolution} strategy. The key intuition is that tools can usually be refined based on use experience. Meanwhile, we also notice that, in our framework, the MLLM's tool-augmented reasoning trajectories provide an observable form of such use experience. Specifically, during RL optimization, each trajectory records which tools are invoked and on which samples they are used. By comparing the final verdict in each trajectory with the ground-truth label, we can further determine whether the detection succeeds. Therefore, these trajectories can be treated as the MLLM's accumulated \textit{tool-use experience} and used as feedback to \ul{evaluate} and \ul{update} the toolset.

\noindent\ul{User-experience-guided tool evaluation.} To evaluate tools from the MLLM's tool-use experience, inspired by how tools are commonly judged in real-world usage~\cite{bevan1994usability,bevan1995measuring}, we consider two complementary aspects: whether a tool is broadly needed and whether it is effective when used. Accordingly, for each tool \(\mathfrak t\), we define its \textit{usage breadth} and \textit{evidential effectiveness}. 

Specifically, \textit{usage breadth} measures whether a tool is broadly needed by the current MLLM. It can be formally defined as the fraction of tool-use trajectories in which the tool is invoked:
$
U(\mathfrak t) =
\frac{1}{|\mathcal{Z}_{\mathrm{buf}}|}
\sum_{j=1}^{|\mathcal{Z}_{\mathrm{buf}}|}
\mathds{1}[\mathfrak t \in \tau_j],
$
where \(\mathcal{Z}_{\mathrm{buf}}\) denotes the set of all tool-use trajectories produced during the previous model-optimization step, and \(\tau_j\) denotes the \(j\)-th trajectory. A higher usage breadth indicates that the MLLM frequently considers this type of evidence relevant across different videos. Meanwhile, \textit{evidential effectiveness} measures whether a tool tends to support correct detection once it is used. It can be defined as the fraction of correct detections among the trajectories that invoke this tool:
$
S(\mathfrak t) =
\frac{
\sum_{j=1}^{|\mathcal{Z}_{\mathrm{buf}}|}
\mathds{1}[\mathfrak t \in \tau_j]
\mathds{1}[\hat{y}_j = y_j]
}{
\sum_{j=1}^{|\mathcal{Z}_{\mathrm{buf}}|}
\mathds{1}[\mathfrak t \in \tau_j]
}.
$
Here, \(\hat{y}_j\) and \(y_j\) denote the predicted and ground-truth labels of the video associated with trajectory \(\tau_j\), respectively. A higher evidential effectiveness indicates that, when the tool is used, it is more often associated with correct detection.

% \begin{wrapfigure}[6]{r}{0.3\textwidth}
% \vspace{-0.45cm}
% \centering
% \includegraphics[width=0.3\textwidth]{figure_add.png}
% \vspace{-0.5cm}
% \caption{Illustration of four types of tools. \qhx{To edit.}}
% \label{fig:ow}
% \end{wrapfigure}
% 

Based on these two measures, all tools in the current toolset can be divided into four groups. Specifically, we rank all tools by usage breadth and split them into high-breadth and low-breadth halves. Similarly, we rank all tools by evidential effectiveness and split them into high-effectiveness and low-effectiveness halves. Taking the intersections of these two partitions yields four types of tools: (a) \textit{high-breadth and high-effectiveness} tools, (b) \textit{high-breadth but low-effectiveness} tools, (c) \textit{low-breadth but high-effectiveness} tools, and (d) \textit{low-breadth and low-effectiveness} tools. These four types reflect different statuses of tools in the current toolset, which guide the subsequent tool updates.

% Based on these two measures, all tools in the current toolset can be divided into four groups. Specifically, we rank all tools according to usage breadth and split them into a high-breadth half and a low-breadth half. Similarly, we rank all tools according to evidential effectiveness and split them into a high-effectiveness half and a low-effectiveness half. Taking the intersections of these two partitions gives four types of tools: high-breadth and high-effectiveness tools, high-breadth but low-effectiveness tools, low-breadth but high-effectiveness tools, and low-breadth and low-effectiveness tools. These four types reflect different statuses of tools in the current toolset, which guide their subsequent tool updates.

% 沾一下进化算法

\ul{User-experience-guided tool update.}
At this point, from the perspective of the MLLM's tool-use experience, we have obtained an evaluation (diagnosis) of each tool. The remaining problem is how to use this diagnosis to effectively update the toolset. This is however still non-trivial because the evaluation only reveals the current status of each tool, but does not directly specify how a custom-formatted tool should be modified.

A natural first step is to make direct decisions for the most certain cases: tools with both high breadth and high effectiveness can be kept, while tools with both low breadth and low effectiveness can be removed. However, this only provides a partial solution. First, it does not specify how to handle partially useful tools, such as tools that are frequently needed but currently unreliable, or tools that are effective but only applicable to limited scenarios. Second, it only exploits the current toolset and provides no mechanism for exploring potentially better tools beyond the existing design space.

To tackle this problem, inspired by model--tool co-evolution~\cite{toolfg2026}, biological evolution, and Genetic Programming~\cite{fisher1999genetical,koza1994}, we design a three-step tool-use-experience-guided tool evolution procedure. The key idea is to convert different diagnostic outcomes into different evolutionary operations: selection for clearly useful or unhelpful tools, mutation for partially useful tools, and exploration for missing tool capabilities. \textit{(1) Selection.} Following the natural first step and mimicking natural selection, we first retain tools with both high breadth and high effectiveness, while pruning tools with both low breadth and low effectiveness. \textit{(2) Mutation.} For partially useful tools, we perform targeted mutation by prompting the offline teacher MLLM to modify their callable Python functions according to their diagnosed limitations. Analogous to genetic mutation, this step introduces local variants of existing tools instead of directly discarding them. Specifically, high-breadth but low-effectiveness tools are mutated toward refinement, since they are frequently needed but not yet reliable; low-breadth but high-effectiveness tools are mutated toward generalization, since they are useful when invoked but may be too narrowly applicable. \textit{(3) Exploration.} Selection and mutation still operate mainly within the current toolset. To explore potentially useful tools beyond the existing design space, we introduce new candidate tools through crossover and brainstorming. Analogous to genetic recombination, crossover combines complementary tools to produce new tools that inherit useful capabilities from both. Brainstorming further introduces missing capabilities by generating tools from failure cases or under-covered unnatural artifacts. Both operations use the same offline teacher interface, which generates Python functions as individual tools or as wrappers that invoke pretrained visual backends. Before admission, every candidate tool is trial-executed on calibration examples to ensure runtime stability: it must import successfully, run without an uncaught exception, and emit the required XML-style schema. Failed candidates are repaired or rejected. Candidates that pass these checks are added to the toolset, used in the next round of model optimization, and can later be pruned if they have low usage breadth and low evidential effectiveness. The \hyperref[app:evo_execution]{trial-execution details} and \hyperref[app:evo_cases]{representative mutation, brainstorming, and crossover cases} are provided in Appendix~\ref{app:evolution}.

\begingroup\looseness=-1
Through the above process, our proposed user-experience-guided tool update strategy transforms tool optimization from an ill-defined and non-trivial problem over custom tools into a tractable feedback-driven tool evolution process. By alternating model optimization and tool evolution, our framework progressively improves both the MLLM's tool-use capability and the suitability of the toolset, ultimately enabling more reliable evidence-grounded AI-generated video detection.
\par\endgroup

%-----------------------------------------------------------------------------
\subsection{Overall Training and Testing}
\label{sec:inference}

\textbf{Training.}
% Our training follows a two-stage process, as illustrated in Fig.~\ref{fig:overview}. Given a general MLLM, the first stage (introduced in Sec.~\ref{sec:stage1}) constructs an initial heuristic toolset and trains the model to acquire tool-use ability. The second stage (introduced in Sec.~\ref{sec:stage2}) alternates between model update and tool update. Specifically, after every $\Delta_{\mathrm{evo}}$ model update steps, we evolve (update) the toolset using the collected tool-use experience. This process continues until training finishes.
\begingroup\looseness=-1
Our training follows a two-stage process, as illustrated in Fig.~\ref{fig:overview}. Given a general MLLM, the first stage constructs an initial heuristic toolset and trains the model to acquire tool-use ability through the influence-based teacher-knowledge guidance strategy introduced in Sec.~\ref{sec:stage1}. The second stage alternates between RL-based model optimization and user-experience-guided tool evolution introduced in Sec.~\ref{sec:stage2}. Specifically, after every $\Delta_{\mathrm{evo}}$ RL optimization steps, we evolve the toolset using the collected tool-use experience. This process continues until training finishes. 
\par\endgroup
%yielding a tool-using expert model equipped with a high-quality toolset.

\textbf{Testing.}
At test time, the trained model adaptively invokes suitable tools, interprets their returned evidence, and reasons over potentially conflicting observations before producing the final verdict. Tool-use statistics and inference details are provided in Appendix~\ref{app:testtime}.

%=============================================================================
\section{Experiments}
\label{sec:exp}
%=============================================================================

% Main-paper result and ablation tables.
% Declared at the start of the Experiments section so two-column floats
% can be placed before the references within the eight-page body limit.

\begin{table*}[t]
\centering
\caption{Performance on ViF-Bench, whose paired splits contain real and fake videos. Ours attains 98.82\% accuracy on the real videos.}
\label{tab:vifbench}
% \vspace{1mm}
\renewcommand{\arraystretch}{1.0}
\resizebox{0.80\textwidth}{!}{%
\begin{tabular}{c|c|ccccccccccccccccccc|c}
\toprule
Method &
\makecell{Metric} &
\makecell{Wan2.1\\-1.3B} &
\makecell{CogV\\-X1.5} &
\makecell{Wan2.2\\-5B\\(T2V)} &
\makecell{Wan2.2\\-5B\\(I2V)} &
\makecell{Hunyuan\\Video\\(T2V)} &
\makecell{Hunyuan\\Video\\(I2V)} &
\makecell{VACE\\-1.3B} &
\makecell{Wan2.2\\-14B\\(T2V)} &
\makecell{Wan2.2\\-14B\\(I2V)} &
\makecell{Skyreels\\-V2\\(T2V)} &
\makecell{Skyreels\\-V2\\(I2V)} &
\makecell{LTX-Video\\-13B\\(T2V)} &
\makecell{LTX-Video\\-13B\\(I2V)} &
\makecell{Gen4\\-Turbo} &
\makecell{Hai-\\luo-02} &
\makecell{Pika\\-V2} &
\makecell{Pixverse\\-V4-5} &
\makecell{Kling\\-V1} &
Sora-2 &
Avg
\\
\midrule
\multirow{3}{*}{BusterX++~\cite{busterx2025}}
  & Acc
  & 54.85 & 59.39 & 52.42 & 50.30 & 59.15 & 49.70 & 50.91 & 62.42 & 49.70 & 65.76 & 50.00 & 56.25 & 50.00 & 50.89 & 61.68 & 76.82 & 75.66 & 52.84 & 52.33 & 56.90
    \\
  & R
  & 10.30 & 19.39 & 5.45 & 1.21 & 18.90 & 0.00 & 2.42 & 25.45 & 0.00 & 32.12 & 0.61 & 13.12 & 0.61 & 2.68 & 24.09 & 54.30 & 51.97 & 5.67 & 5.33 & 14.40
    \\
  & F1
  & 18.58 & 32.32 & 10.29 & 2.38 & 31.63 & 0.00 & 4.71 & 40.38 & 0.00 & 48.40 & 1.20 & 23.08 & 1.21 & 5.17 & 38.60 & 70.09 & 68.10 & 10.74 & 10.06 & 21.94
    \\
 \midrule
\multirow{3}{*}{\makecell{VidGuard-\\R1~\cite{vidguard2025}}}
  & Acc
  & 97.55 & 94.79 & 90.18 & 84.05 & 93.83 & 76.38 & 73.93 & 95.71 & 81.60 & 95.40 & 79.01 & 96.88 & 82.82 & 77.48 & 94.12 & 96.67 & 97.00 & 94.64 & 88.18 & 88.96
    \\
  & R
  & 98.77 & 93.25 & 84.05 & 71.78 & 91.36 & 56.44 & 51.53 & 95.09 & 66.87 & 94.48 & 61.73 & 97.50 & 69.33 & 60.36 & 92.65 & 97.33 & 98.00 & 93.57 & 80.41 & 81.82
    \\
  & F1
  & 97.58 & 94.70 & 89.54 & 81.82 & 93.67 & 70.50 & 66.40 & 95.68 & 78.42 & 95.36 & 74.63 & 96.89 & 80.14 & 72.83 & 94.03 & 96.69 & 97.03 & 94.58 & 87.18 & 87.25
    \\
 \midrule
\multirow{3}{*}{\makecell{Video-\\Veritas~\cite{videoversitas2026}}}
  & Acc
  & 98.16 & 97.85 & 93.56 & 86.81 & 94.75 & 69.33 & 75.15 & 96.93 & 86.20 & 96.93 & 81.48 & 97.81 & 85.28 & 81.08 & 97.06 & 98.00 & 97.67 & 97.50 & 90.54 & 90.64
    \\
  & R
  & 100.00 & 99.39 & 90.80 & 77.30 & 93.21 & 42.33 & 53.99 & 97.55 & 76.07 & 97.55 & 66.67 & 99.38 & 74.23 & 66.67 & 98.53 & 100.00 & 99.33 & 99.29 & 84.46 & 85.09
    \\
  & F1
  & 98.19 & 97.89 & 93.38 & 85.42 & 94.67 & 57.98 & 68.48 & 96.95 & 84.64 & 96.95 & 78.26 & 97.85 & 83.45 & 77.89 & 97.10 & 98.04 & 97.70 & 97.54 & 89.93 & 89.07
    \\
 \midrule
\multirow{3}{*}{\makecell{Ivy-\\xDetector~\cite{ivyfake2025}}}
  & Acc
  & 96.93 & 88.96 & 89.26 & 83.74 & 94.14 & 65.64 & 73.31 & 96.01 & 79.75 & 95.09 & 76.85 & 96.56 & 80.98 & 74.77 & 96.32 & 98.00 & 97.00 & 95.36 & 87.50 & 87.69
    \\
  & R
  & 98.16 & 82.21 & 82.82 & 71.78 & 92.59 & 35.58 & 50.92 & 96.32 & 63.80 & 94.48 & 58.02 & 97.50 & 66.26 & 55.86 & 97.79 & 100.00 & 98.67 & 95.71 & 79.73 & 79.91
    \\
  & F1
  & 96.97 & 88.16 & 88.52 & 81.53 & 94.04 & 50.88 & 65.61 & 96.02 & 75.91 & 95.06 & 71.48 & 96.59 & 77.70 & 68.89 & 96.38 & 98.04 & 97.05 & 95.37 & 86.45 & 85.30
    \\
 \midrule
\multirow{3}{*}{Skyra~\cite{skyra2026}}
  & Acc
  & 96.97 & 96.36 & 92.12 & 87.58 & 94.82 & 93.64 & 79.09 & 96.36 & 84.55 & 95.76 & 78.96 & 95.94 & 83.74 & 79.46 & 95.99 & 96.36 & 96.05 & 94.68 & 91.00 & 91.02
    \\
  & R
  & 100.00 & 98.79 & 90.30 & 81.21 & 95.73 & 93.33 & 64.24 & 98.79 & 75.15 & 97.58 & 64.02 & 98.12 & 73.62 & 66.07 & 98.54 & 99.34 & 98.68 & 96.45 & 88.67 & 88.35
    \\
  & F1
  & 97.06 & 96.45 & 91.98 & 86.73 & 94.86 & 93.62 & 75.44 & 96.45 & 82.94 & 95.83 & 75.27 & 96.02 & 81.91 & 76.29 & 96.09 & 96.46 & 96.15 & 94.77 & 90.78 & 90.27
    \\
  \midrule
\multirow{3}{*}{Ours}
  & Acc
  & 99.39 & 99.39 & 96.63 & 94.79 & 98.77 & 95.09 & 92.94 & 99.08 & 94.17 & 99.08 & 93.21 & 99.06 & 96.32 & 91.44 & 99.26 & 99.00 & 99.34 & 98.57 & 95.61 & \textbf{96.90}
    \\
  & R
  & 100.00 & 100.00 & 94.48 & 90.80 & 98.77 & 91.41 & 87.12 & 99.39 & 89.57 & 99.39 & 87.65 & 99.38 & 93.87 & 83.78 & 100.00 & 99.34 & 99.34 & 97.86 & 92.57 & \textbf{94.98}
    \\
  & F1
  & 99.39 & 99.39 & 96.55 & 94.57 & 98.77 & 94.90 & 92.51 & 99.08 & 93.89 & 99.08 & 92.81 & 99.07 & 96.23 & 90.73 & 99.27 & 99.00 & 99.34 & 98.56 & 95.47 & \textbf{96.77}
    \\
\bottomrule
\end{tabular}%
}
\vspace{-0.3cm}
\end{table*}

\begin{table*}[t]
\centering
%\renewcommand{\arraystretch}{1.05}
% \vspace{-0.2cm}
\caption{Performance on GenVideo, whose test splits combine generated videos with a shared real-video pool. Ours' near-perfect R and F1 require strong performance on both classes.}
\label{tab:genvideo}
\resizebox{0.56\textwidth}{!}{
\begin{tabular}{c|c|cccccccccc|c}
\toprule
\multirow{2}{*}{Method} & \multirow{2}{*}{Metric}&\multirow{2}{*}{Sora}&Morph&\multirow{2}{*}{Gen2}&\multirow{2}{*}{HotShot}&\multirow{2}{*}{Lavie}&\multirow{2}{*}{Show-1}&Moon&\multirow{2}{*}{Crafter}&Model&Wild&\multirow{2}{*}{Avg}\\
&&&Studio&&&&&Valley&&Scope&Scrape&\\
\midrule
\multirow{2}{*}{NPR~\cite{npr2024}} &
R  & 0.91 & 0.99 & 0.99 & 0.24 & 0.89 & 0.57 & 0.97 & 0.99 & 0.94 & 0.87 & 0.84 \\
& F1 & 0.27 & 0.84 & 0.91 & 0.30 & 0.86 & 0.59 & 0.81 & 0.91 & 0.81 & 0.81 & 0.71 \\
\midrule
\multirow{2}{*}{VideoMAE~\cite{videomae2022}} &
R  & 0.67 & 0.96 & 0.98 & 0.96 & 0.77 & 0.80 & 0.97 & 0.96 & 0.96 & 0.68 & 0.87 \\
& F1 & 0.62 & 0.95 & 0.98 & 0.96 & 0.86 & 0.87 & 0.96 & 0.97 & 0.96 & 0.79 & 0.89 \\
\midrule
\multirow{2}{*}{MINTIME-CLIP~\cite{mintime2024}} &
R  & 0.89 & 1.00 & 0.98 & 0.26 & 0.96 & 0.98 & 0.99 & 1.00 & 0.84 & 0.82 & 0.87 \\
& F1 & 0.49 & 0.93 & 0.96 & 0.37 & 0.94 & 0.92 & 0.92 & 0.96 & 0.84 & 0.85 & 0.82 \\
\midrule
\multirow{2}{*}{FTCN-CLIP~\cite{ftcn2021}} &
R  & 0.87 & 1.00 & 0.98 & 0.17 & 0.97 & 0.91 & 1.00 & 1.00 & 0.85 & 0.82 & 0.86 \\
& F1 & 0.78 & 0.98 & 0.98 & 0.29 & 0.98 & 0.94 & 0.98 & 0.99 & 0.90 & 0.89 & 0.87 \\
\midrule
\multirow{2}{*}{DeMamba-XCLIP~\cite{genvideo2024}} &
R  & 0.98 & 1.00 & 0.99 & 0.65 & 0.94 & 0.98 & 1.00 & 1.00 & 0.92 & 0.89 & 0.93 \\
& F1 & 0.64 & 0.96 & 0.97 & 0.75 & 0.95 & 0.95 & 0.95 & 0.97 & 0.92 & 0.91 & 0.90 \\
\midrule
\multirow{2}{*}{VidGuard-R1~\cite{vidguard2025}} &
R  & 0.95 & 1.00 & 0.98 & 0.94 & 0.98 & 0.95 & 0.97 & 0.99 & 0.94 & 0.91 & 0.96 \\
& F1 & 0.97 & 0.99 & 0.99 & 0.91 & 0.99 & 0.89 & 0.99 & 0.99 & 0.95 & 0.90 & 0.96 \\
\midrule
\multirow{2}{*}{Skyra~\cite{skyra2026}} &
R  & 0.68 & 0.82 & 0.88 & 0.70 & 0.81 & 0.78 & 0.96 & 0.90 & 0.79 & 0.83 & 0.81 \\
& F1 & 0.81 & 0.90 & 0.94 & 0.82 & 0.90 & 0.88 & 0.98 & 0.95 & 0.88 & 0.91 & 0.90 \\
\midrule
\multirow{2}{*}{Ours} &
R  & 1.00 & 1.00 & 1.00 & 1.00 & 0.99 & 1.00 & 1.00 & 1.00 & 0.99 & 1.00 & \textbf{1.00} \\
& F1 & 1.00 & 1.00 & 1.00 & 1.00 & 0.99 & 1.00 & 1.00 & 1.00 & 0.99 & 1.00 & \textbf{1.00} \\
\bottomrule
\end{tabular}}
\vspace{-0.5cm}
\end{table*}

\begin{table}[t]
\centering
\caption{Evaluation on the key strategies in our framework.}
\label{tab:abl_components}
\resizebox{0.78\linewidth}{!}
{
\small
\begin{tabular}{lccc}
\hline
Method & Acc & R & F1 \\
\hline
w/o influence-based guidance               & 95.70 & 91.94 & 95.34 \\
w/o user-experience-guided tool evolution                  & 92.92 & 93.98 & 92.91 \\
\hline
TUE-Detector                   & \textbf{96.90} & \textbf{94.98} & \textbf{96.77} \\
\hline
\end{tabular}}
\end{table}

\begin{table}[t]
\centering
\caption{Evaluation on the two-stage training process.}
\label{tab:abl_stages}
{
\small
\begin{tabular}{lccc}
\hline
Method & Acc & R & F1 \\
\hline
w/o Stage~1                          & 82.99 & 70.71 & 79.39  \\
w/o Stage~2                          & 93.12 & 93.85 & 93.08  \\
\hline
TUE-Detector                    & \textbf{96.90} & \textbf{94.98} & \textbf{96.77} \\
\hline
\end{tabular}}
\end{table}

\noindent\textbf{Datasets and evaluation metrics.}
To evaluate TUE-Detector, we conduct experiments on ViF-Bench~\cite{skyra2026} and GenVideo~\cite{genvideo2024}.
GenVideo is a widely-used dataset for AI-generated video detection introduced in the DeMamba work~\cite{genvideo2024}.
Following~\cite{vidguard2025}, we use the many-to-many zero-shot protocol on GenVideo and report Recall (R) and F1-score (F1). ViF-Bench~\cite{skyra2026} is a more recent and comprehensive dataset. For this dataset, we follow Skyra~\cite{skyra2026} and report accuracy (ACC), R, and F1. More details are in Appendix~\ref{app:data_metrics}. Our current evaluation focuses on fully AI-generated text-to-video and image-to-video videos. The proposed framework could also be adapted to face-manipulation detection by incorporating task-specific forensic tools, and we leave its evaluation on benchmarks such as FaceForensics++ to future work.

\noindent\textbf{Implementation details.}
We conduct the main experiments on 8 NVIDIA H200 GPUs, and the complete TUE-Detector training run takes 19.3 hours.
We initialize the detector from Qwen2.5-VL-7B-Instruct~\cite{qwen25vl2025}.
We use a capable API-accessible MLLM as an offline teacher/proposal model, which is never called at inference time. Teacher reliability and reconstructed API cost are reported in Appendix~\ref{app:teacher_reliability_cost}; teacher-swap and tool-registry reproducibility details are in Appendices~\ref{app:teacher_model} and~\ref{app:evo_snapshot}.

\subsection{Main Results}
\label{sec:main_results}

\begingroup\looseness=-1
We compare with state-of-the-art AI-generated video detection methods. Experimental results on ViF-Bench and GenVideo are reported in Tab.~\ref{tab:vifbench} and Tab.~\ref{tab:genvideo}. On ViF-Bench, TUE-Detector achieves the best average performance across generators. On GenVideo, it reaches near-saturated performance under the established many-to-many zero-shot protocol. We therefore use ViF-Bench as the primary discriminative benchmark and retain GenVideo mainly for cross-generator comparability with prior work.
\par\endgroup

\subsection{Ablation Studies}
\label{sec:ablation}

We conduct extensive ablation experiments on ViF-Bench and report the average performance across generators. \textbf{Additional ablation studies are provided in Appendix~\ref{app:moreablations}.}

\begingroup\looseness=-1
\textbf{Impact of the key strategies in our framework.} In our TUE-Detector framework, we introduce two key strategies: the influence-based teacher-knowledge guidance strategy in training stage 1 and the user-experience-guided tool evolution strategy in training stage 2. To assess the contribution of each strategy, we test two variants. In the first variant (\textbf{w/o influence-based guidance}), we discard the influence-based teacher-knowledge guidance strategy, and directly use Eq.~\ref{eq:naive_sft} for stage 1 training. In the second variant (\textbf{w/o user-experience-guided tool evolution}), we discard the user-experience-guided tool evolution strategy and do not optimize tool in stage 2 training. As shown in Tab.~\ref{tab:abl_components}, compared to our framework, the performance of both variants drops significantly, showing the efficacy of both key strategies in our framework.
\par\endgroup

\textbf{Impact of the two-stage training process.} In our framework, training proceeds through a progressive two-stage process. To evaluate the effectiveness of this design, we consider two variants. In the first variant (\textbf{w/o Stage~1}), we remove the training process in training stage 1 and, after initializing the heuristic toolset, directly optimize the model and toolset alternately as in training stage 2. In the second variant (\textbf{w/o Stage~2}), we omit training stage 2 and directly evaluate the tool-usage expert detector obtained after training stage 1. As shown in Tab.~\ref{tab:abl_stages}, our framework significantly outperforms both variants, demonstrating the importance of both stages in our training process. Both directly prompting off-the-shelf models with the updated toolset and omitting the task-specific tool-use and evidence-integration training of Stage~1 lead to lower performance. Additional variants are provided in Appendix~\ref{app:impact_training}.

%=============================================================================
\section{Conclusion}
\label{sec:conclusion}
%=============================================================================

\begingroup\looseness=-1
In this paper, we proposed TUE-Detector, a novel framework for AI-generated video detection from the perspective of tool-mediated evidence discovery. TUE-Detector enables training a general MLLM into a task-tailored, tool-using expert detector for AI-generated video detection. In TUE-Detector, we propose several novel designs to equip the detector with high-quality, suitable tools and strong tool-use capability. Extensive experiments demonstrate the efficacy of TUE-Detector.
\par\endgroup

%=============================================================================
% References
%=============================================================================
{\small
\nocite{gpt54openai2026,openaiimagetokens2026,grpo2024}
\bibliographystyle{ieeenat_fullname}
\bibliography{references}
}

\clearpage
% Continue figure, table, and equation numbering from the main paper.
\setcounter{figure}{2}
\setcounter{table}{4}
\setcounter{equation}{5}
\appendix
\section*{Supplementary Material}
%=============================================================================

This appendix is organised as follows. Section~\ref{app:supp_exp} presents additional ablation studies. Section~\ref{app:supp_vis} presents additional visualisations and case studies. Section~\ref{app:additional_details} provides additional experimental details and the mathematical derivation of the influence reweighting stage. Section~\ref{app:toolset_details} provides toolset details. Section~\ref{app:supp_prompts} collects the prompts referenced in the paper. Section~\ref{app:licenses} contains the dataset and model licenses.

\section{Additional Ablation Studies}
\label{app:supp_exp}
\label{app:B_ablations}
\label{app:moreablations}
%=============================================================================

In this section, we present the additional ablation studies promised in Section~\ref{sec:ablation} of the main paper. Unless noted otherwise, we conduct experiments on VIF-Bench and report the average performance across generators.

\noindent\textbf{Impact of the two complementary aspects of tool-use experience.}\label{app:ablation_dimensions}
In our framework, we introduce two complementary aspects of tool-use experience to evaluate tools, namely $U(\mathfrak t)$ measuring the usage breadth of a tool across rollouts and $S(\mathfrak t)$ measuring the evidential effectiveness of a tool when it is invoked. To evaluate the efficacy of each aspect, we test two variants. In the first variant (\textbf{w/o $U(\mathfrak t)$}), we remove the usage-breadth aspect from tool evaluation. In the second variant (\textbf{w/o $S(\mathfrak t)$}), we remove the evidential-effectiveness aspect from tool evaluation. For these two variants, since each only has one evaluation aspect, we keep high-breadth or high-effectiveness tools, while for tools that is determined as low-breadth or low-effectiveness, we let the MLLM decide whether to prune the tool from the toolset or to mutate it. Moreover, for both variants, brainstorming and crossover operations are kept. As shown in Table~\ref{tab:ablation_dimensions}, compared to our framework, the performance of each variant drops, which shows the importance of evaluating tools along both complementary aspects.

\begin{table}[H]
\centering
\caption{Evaluation on the two complementary aspects of tool-use experience.}
\label{tab:ablation_dimensions}
\small
\begin{tabular}{l|ccc}
\hline
Method & Acc\,$\uparrow$ & R\,$\uparrow$ & F1\,$\uparrow$ \\
\hline
w/o $U(\mathfrak t)$ & 95.58 & 93.48 & 95.35 \\
w/o $S(\mathfrak t)$ & 95.27 & 93.52 & 95.06 \\
TUE-Detector         & 96.90 & 94.98 & 96.77 \\
\hline
\end{tabular}
\end{table}

\noindent\textbf{Impact of the toolset update strategy.}\label{app:evo_naive_full}
In our framework, the toolset is evolved together with the model policy through selection, mutation, and exploration, where exploration includes brainstorming and crossover. As mentioned in Section~\ref{sec:stage2} with the phrase \emph{partially useful tools}, here we still use this term to collectively refer to high-breadth but low-effectiveness and low-breadth but high-effectiveness tools. Here, to test the efficacy of our toolset update strategy, we test a variant. In this we remove the toolset update strategy. Ablation drops below \method{}, which shows that brainstorming, and crossover all contribute to the final detector.

\begin{table}[H]
\centering
\caption{Evaluation on the toolset update strategy.}
\label{tab:naive_vs_full_evo}
\small
\begin{tabular}{p{0.48\linewidth}|ccc}
\hline
Configuration & Acc\,$\uparrow$ & R\,$\uparrow$ & F1\,$\uparrow$ \\
\hline
w/o tool update strategy                & 93.79 & 91.36 & 93.37 \\
TUE-Detector                            & 96.90 & 94.98 & 96.77 \\
\hline
\end{tabular}
\end{table}

\noindent\textbf{Impact of training the model to use tools.}\label{app:impact_training}
In our framework, training is used to turn a general MLLM into a task-tailored tool-using detector. To evaluate the impact of this design choice, we compare our framework with variants that directly use off-the-shelf large models without training and prompt Qwen2.5-VL-7B-Instruct or GPT-5.4 to use the initial or updated toolset. As shown in Table~\ref{tab:impact_training}, these variants suffer a substantial drop even using the updated toolset, while \method{} achieves much stronger performance. This shows that the toolset alone is insufficient: our way of model training is critical for learning when to invoke tools, how to integrate their evidence, and how to make reliable AI-generated video detection decisions.

\begin{table}[H]
\centering
\caption{Evaluation on training the model to use tools.}
\label{tab:impact_training}
\small
\setlength{\tabcolsep}{3.5pt}
\begin{tabular}{p{0.48\linewidth}|ccc}
\hline
Model & Acc\,$\uparrow$ & R\,$\uparrow$ & F1\,$\uparrow$ \\
\hline
Qwen2.5-VL-7B-Instruct (initial toolset) & 65.33 & 44.15 & 52.94 \\
Qwen2.5-VL-7B-Instruct (updated toolset) & 70.58 & 44.13 & 57.68 \\
GPT-5.4 (initial toolset) & 69.13 & 59.44 & 64.73 \\
GPT-5.4 (updated toolset) & 75.06 & 50.77 & 65.72 \\
TUE-Detector & 96.90 & 94.98 & 96.77 \\
\hline
\end{tabular}
\end{table}

\noindent\textbf{Tool-mediated reasoning versus learned aggregation.}\label{app:output_aggregation}
In our framework, the MLLM adaptively invokes tools, interprets their returned evidence, and integrates that evidence into its reasoning throughout the detection trajectory. To evaluate whether the performance gain depends on this tool-mediated reasoning process or can be reproduced by learned aggregation alone, we test a variant. In this variant (\textbf{learned aggregation}), we train a simple MLP to combine Skyra's final prediction with the outputs of our final toolset, without the adaptive tool invocation and evidence reasoning used in our framework. As shown in Table~\ref{tab:output_aggregation}, this variant remains substantially below TUE-Detector across all three metrics. These results show that learned aggregation alone is insufficient and highlight the benefit of adaptive tool invocation and evidence reasoning within the detection trajectory.

\begin{table}[H]
\centering
\caption{Comparison with learned aggregation on ViF-Bench.}
\label{tab:output_aggregation}
\small
\setlength{\tabcolsep}{5pt}
\begin{tabular}{p{0.48\linewidth}|ccc}
\hline
Method & Acc\,$\uparrow$ & R\,$\uparrow$ & F1\,$\uparrow$ \\
\hline
Learned aggregation (MLP) & 91.12 & 82.89 & 88.55 \\
TUE-Detector          & 96.90 & 94.98 & 96.77 \\
\hline
\end{tabular}
\end{table}

\noindent\textbf{Impact of teacher model used for trajectory generation.}\label{app:teacher_model}
In our framework, when constructing the tool augmented trajectories during Stage~1 and evolving the toolset during Stage~2, we use GPT-5.4 as the teacher model. Here we also assess other choices of teacher model, including Claude Opus 4.7 and Gemini 3.1 Pro, and report the results in Table~\ref{tab:teacher_model}. As shown in Table~\ref{tab:teacher_model}, with different choices of teacher model, our framework performs consistently, which shows the robustness of our framework to the choice of teacher model.

\begin{table}[H]
\centering
\caption{Evaluation on teacher model used for trajectory generation.}
\label{tab:teacher_model}
\small
\setlength{\tabcolsep}{6pt}
\begin{tabular}{p{0.48\linewidth}|ccc}
\hline
Teacher model & Acc\,$\uparrow$ & R\,$\uparrow$ & F1\,$\uparrow$ \\
\hline
Claude Opus 4.7 & 97.15 & 94.76 & 96.91 \\
Gemini 3.1 Pro & 96.91 & 94.89 & 96.66 \\
GPT-5.4 & 96.90 & 94.98 & 96.77 \\
\hline
\end{tabular}
\end{table}

\noindent\textbf{Impact of exposing ground-truth binary label to the teacher model when constructing the tool augmented reference trajectories.}\label{app:teacher_gt}
In our framework (\textbf{w/o GT binary label}), during prompting the teacher model to generate trajectories, we do not pass it with the binary ground-truth label. To evaluate this design, we test a variant (\textbf{with GT binary label}), in which the GT binary label is instead exposed to the teacher MLLM during its trajectory generation. As shown in Table~\ref{tab:teacher_gt}, exposing the GT label to the teacher leads to a drop across all three metrics. A potential reason is that, once the teacher already knows the answer, it can fall back on the binary label as a shortcut and become less invested in carefully invoking tools, interpreting their outputs, and assembling evidence into a well-grounded reasoning chain, slightly degrading the quality of tool-grounded reasoning that the student inherits from the trajectories.

\begin{table}[H]
\centering
\caption{Evaluation on exposing ground-truth binary label to the teacher model when constructing the tool augmented reference trajectories.}
\label{tab:teacher_gt}
\small
\setlength{\tabcolsep}{6pt}
\begin{tabular}{p{0.48\linewidth}|ccc}
\hline
Teacher input & Acc\,$\uparrow$ & R\,$\uparrow$ & F1\,$\uparrow$ \\
\hline
with GT binary label              & 95.70 & 93.23 & 95.45 \\
w/o GT binary label               & 96.90 & 94.98 & 96.77 \\
\hline
\end{tabular}
\end{table}

\noindent\textbf{Impact of our token-level mask of tool output.}\label{app:token_level_mask_tool_output}
In our framework, we introduce the token-level mask of tool output during training in Stage~1, which let the model focus learning on tool-use decisions and evidence-based reasoning, rather than also learning tool output. To evaluate the efficacy of this mask, we test a variant. In this variant (\textbf{w/o token-level mask of tool output}), we train the model in Stage~1 without masking the token-level mask of tool output, while keeping all other components of our framework fixed. As shown in Table~\ref{tab:token_level_mask_tool_output}, compared to our framework, the performance of this variant drops, which shows the importance of this token-level mask of tool output during training in Stage~1.

\begin{table}[H]
\centering
\caption{Evaluation on our token-level mask of tool output.}
\label{tab:token_level_mask_tool_output}
\small
\begin{tabular}{p{0.48\linewidth}|ccc}
\hline
Method & Acc\,$\uparrow$ & Recall\,$\uparrow$ & F1\,$\uparrow$ \\
\hline
w/o token-level mask of tool output & 89.75 & 89.60 & 89.47 \\
TUE-Detector                         & 96.90 & 94.98 & 96.77 \\
\hline
\end{tabular}
\end{table}

\noindent\textbf{LLM-judge evaluation of detection trajectories.}\label{app:llm_judge_trajectory_quality}
We further evaluate the quality of different methods' detection reasoning using Claude Opus 4.7 as an external LLM judge. The judge rates how well each method captures \emph{subtle yet measurable unnatural artifacts} on a Likert scale from $1$ to $5$. Its prompt is (Given 16 frames, timestamps, GT, and one CoT, score from 1 to 5 on how well the CoT captures visible, subtle, measurable unnatural artifacts. Penalize missed artifacts, vague claims, contradictions, or hallucinations. For GT=Real, reward avoiding false artifact claims). We compare \method{} against recent methods including Skyra-RL and VidGuard-R1. For each of the three models, we evaluate detection reasoning on the same randomly selected set of $250$ video samples. The evaluated method identity and the video generator name are hidden from the judge. As shown in Table~\ref{tab:llm_judge_trajectory_quality}, our method scores substantially higher than both baselines.

\begin{table}[H]
\centering
\caption{External LLM-judge evaluation of detection reasoning trajectories on a scale from $1$ to $5$ (higher is better). Each method is evaluated on the same set of $250$ video samples.}
\label{tab:llm_judge_trajectory_quality}
\small
\begin{tabular}{l|c}
\hline
Method & LLM Judge\,$\uparrow$ \\
\hline
Skyra-RL                  & 2.0 \\
VidGuard-R1               & 2.1 \\
TUE-Detector              & 4.8 \\
\hline
\end{tabular}
\end{table}

\noindent\textbf{Impact of the adaptive damping coefficient $\kappa$.}\label{app:ablation_kappa}
In our framework, when estimating per sample influence weights, we set the adaptive damping coefficient $\kappa$ in Eq.~\ref{eq:rho_adaptive_app} to be $10$. Here we also assess other choices of $\kappa$ ranging from $1$ to $50$, and report the results in Table~\ref{tab:ablation_kappa}. As shown in Table~\ref{tab:ablation_kappa}, our framework gets optimal performance when $\kappa$ is set to $10$ or $20$, and $\kappa = 10$ is used in our experiments. The damping term is introduced to make the empirical Fisher/ridge system numerically stable by avoiding singular or near zero directions. In this run, changing $\kappa$ mostly rescales the influence magnitudes and preserves the sign/ranking of the samples, so the final metrics vary only slightly. Besides, with different choices of $\kappa$ from $1$ to $50$, our framework outperforms the previous state of the art method consistently. This demonstrates the robustness of our framework to this hyperparameter.

\begin{table}[H]
\centering
\caption{Evaluation on the adaptive damping coefficient $\kappa$.}
\label{tab:ablation_kappa}
\small
\begin{tabular}{c|ccc}
\hline
$\kappa$ & Acc\,$\uparrow$ & R\,$\uparrow$ & F1\,$\uparrow$ \\
\hline
$1$   & $96.35$ & $92.70$ & $96.08$ \\
$5$   & $96.89$ & $94.95$ & $96.75$ \\
$10$  & $96.90$ & $94.98$ & $96.77$ \\
$20$  & $96.90$ & $94.98$ & $96.77$ \\
$50$  & $96.33$ & $92.67$ & $96.06$ \\
\hline
\end{tabular}
\end{table}

\noindent\textbf{Impact of using IF with our ridge reformulation.}
\label{app:if_overhead}
In our framework, we introduce an influence function with a ridge reformulation to practically estimate per-sample training weights. To evaluate this design, we test a variant, \textbf{Standard IF}, which also performs influence-based reweighting but computes influence directly following the standard formulation in Eq.~\ref{eq:inf} in the main paper. As shown in Table~\ref{tab:if_overhead}, our ridge-based computation enables a practically feasible training procedure, whereas the standard variant is computationally intractable and thus impractical.

\begin{table}[H]
\centering
\caption{Evaluation on using IF with our ridge reformulation.}
\label{tab:if_overhead}
\renewcommand{\arraystretch}{1.15}
\setlength{\tabcolsep}{4pt}
\resizebox{0.35\textwidth}{!}{%
\begin{tabular}{lccc}
\toprule
Pipeline & F1 & Total training time \\
\midrule
Standard IF           & N.A. & infeasible\\
TUE-Detector          & 96.77 & 19.3 hours\\
\bottomrule
\end{tabular}%
}
\end{table}

\noindent\textbf{Impact of different ways of weighting.}\label{app:ablation_weighting_alternatives}
To further examine whether the performance gain comes from sample reweighting itself or from the influence function weighting signal, we compare our method with two alternative weighting strategies using the same weighted SFT interface. The variant of our framework (\textbf{weighting with loss}), during its training Stage~1, we set the temperature $\zeta_{\text{loss}}$ to the median loss over the full training set and compute the raw weight as $\widetilde{w}_j = \exp(-\mathcal{L}_j / \zeta_{\text{loss}})$. This weighting rule assigns larger weights to trajectories that the early model can fit more easily, treating them as cleaner and more internally consistent samples, and assigns smaller weights to high loss trajectories that may be harder, noisier, or less consistent. The raw weights are clipped to $[0.10, 3.0]$ and then divided by their training set mean so that the final weights have mean $1$. As shown in Table~\ref{tab:weighting_alternatives}, this alternative underperform our influence function weighting, indicating that the improvement does not come merely from introducing non uniform sample weights but from using a validation targeted influence signal.

\begin{table}[H]
\centering
\caption{Evaluation on different ways of weighting.}
\label{tab:weighting_alternatives}
\small
\setlength{\tabcolsep}{6pt}
\begin{tabular}{l|ccc}
\hline
Weighting strategy & Acc\,$\uparrow$ & Recall\,$\uparrow$ & F1\,$\uparrow$ \\
\hline
weighting with loss & 95.63 & 92.44 & 95.28 \\
TUE-Detector & 96.90 & 94.98 & 96.77 \\
\hline
\end{tabular}
\end{table}

\noindent\textbf{Impact of using influence function or LLM judge weighting.}\label{app:B_stage1}
In our framework, we introduce the influence-based teacher-knowledge guidance strategy, which uses the influence function to assess the quality of the reasoning trajectory generated by the teacher model for each sample, and adjusts the sample's training weight accordingly. To test the efficacy of this design, here we evaluate a variant. In this variant, instead of using our strategy, we use a simple prompt to let GPT-5.4 score the quality of the reasoning trajectory generated by the teacher model for each sample. Specifically, we let GPT-5.4 output an integer score from $1$ to $10$, denoted $s_i$. We then convert the judge score $s_i$ into a sample weight using the same weighted SFT interface as our method $w_i^{\text{judge}} = \frac{s_i}{\frac{1}{J}\sum_{j=1}^{J} s_j}$, where the denominator is the mean score over the training split, so the average sample weight is approximately $1.0$. As shown in Table~\ref{tab:if_vs_llm_judge}, influence-based trajectory reweighting gives substantially stronger performance than the LLM judge weighting baseline.

\begin{table}[H]
\centering
\caption{Evaluation on using influence function or LLM judge weighting.}
\label{tab:if_vs_llm_judge}
\small
\setlength{\tabcolsep}{6pt}
\begin{tabular}{l|ccc}
\hline
Weighting strategy & Acc\,$\uparrow$ & R\,$\uparrow$ & F1\,$\uparrow$ \\
\hline
LLM Judge (GPT-5.4) & 92.58 & 92.12 & 92.37 \\
TUE-Detector        & 96.90 & 94.98 & 96.77 \\
\hline
\end{tabular}
\end{table}

\noindent\textbf{Impact of reward weight $\lambda_{\text{traj}}$.}\label{app:B_reward}\label{app:ablation_alpha}
In Eq.~\ref{eq:reward} in the main paper, our reward has two terms, $\mathcal{R}_{\text{result}}$ and $\mathcal{R}_{\text{trajectory}}$. Here we introduce $\lambda_{\text{traj}}$ to further explore the scale of $\mathcal{R}_{\text{trajectory}}$ relative to $\mathcal{R}_{\text{result}}$, that is, the reward formula becomes $\mathcal{R} = \mathcal{R}_{\text{result}} + \lambda_{\text{traj}} \cdot \mathcal{R}_{\text{trajectory}}$. Here we assess different choices of $\lambda_{\text{traj}}$ ranging from $0.5$ to $1.5$, and report the results in Table~\ref{tab:ablation_alpha}. As shown, with different choices of $\lambda_{\text{traj}}$, our framework performs consistently, which shows the robustness of our framework to $\lambda_{\text{traj}}$.

\begin{table}[H]
\centering
\caption{Evaluation on reward weight $\lambda_{\text{traj}}$.}
\label{tab:ablation_alpha}
\small
\begin{tabular}{c|ccc}
\hline
$\lambda_{\text{traj}}$ & Acc\,$\uparrow$ & R\,$\uparrow$ & F1\,$\uparrow$ \\
\hline
$0.5$  & 96.82 & 94.82 & 96.69 \\
$0.75$ & 96.76 & 94.71 & 96.62 \\
$1$    & 96.90 & 94.98 & 96.77 \\
$1.25$ & 96.84 & 94.85 & 96.70 \\
$1.5$  & 96.86 & 94.91 & 96.73 \\
\hline
\end{tabular}
\end{table}

\noindent\textbf{Impact of each reward component.}\label{app:ablation_reward_component}
In Eq.~\ref{eq:reward} in the main paper, the overall reward combines a direct correctness bonus, $\mathcal{R}_{\text{result}}$, with trajectory-level shaping, $\mathcal{R}_{\text{trajectory}}$. The latter contains both format feedback and the label-dependent anchor term described in Section~\ref{app:reward_trajectory}. We evaluate two ablations: \textbf{w/o $\mathcal{R}_{\text{result}}$} removes only the direct correctness bonus while retaining trajectory-level shaping, whereas \textbf{w/o $\mathcal{R}_{\text{trajectory}}$} removes both the format and anchor terms while retaining the direct correctness bonus. Table~\ref{tab:ablation_reward_component} shows that either removal reduces performance, indicating that the two components provide complementary training signals.

\begin{table}[H]
\centering
\caption{Evaluation of the direct result bonus and trajectory-level shaping reward.}
\label{tab:ablation_reward_component}
\small
\begin{tabular}{l|ccc}
\hline
Method & Acc\,$\uparrow$ & R\,$\uparrow$ & F1\,$\uparrow$ \\
\hline
w/o $\mathcal{R}_{\text{result}}$    & 95.44 & 93.32 & 95.20 \\
w/o $\mathcal{R}_{\text{trajectory}}$    & 95.78 & 93.32 & 95.54 \\
TUE-Detector                     & 96.90 & 94.98 & 96.77 \\
\hline
\end{tabular}
\end{table}

\noindent\textbf{Impact of the tool evolution interval $\Delta_{\mathrm{evo}}$.}\label{app:B_coevo}\label{app:ablation_N}
In our framework, we set the tool evolution interval $\Delta_{\mathrm{evo}}$ to be $20$, meaning that one tool evolution cycle is triggered every $20$ GRPO steps. Here we also assess other choices of $\Delta_{\mathrm{evo}}$ ranging from $5$ to $60$, and report the results in Table~\ref{tab:ablation_N}. As shown, with different choices of $\Delta_{\mathrm{evo}}$, our framework performs consistently, which shows the robustness of our framework to $\Delta_{\mathrm{evo}}$.

\begin{table}[H]
\centering
\caption{Evaluation on the tool evolution interval $\Delta_{\mathrm{evo}}$.}
\label{tab:ablation_N}
\small
\begin{tabular}{c|ccc}
\hline
$\Delta_{\mathrm{evo}}$ (steps) & Acc\,$\uparrow$ & R\,$\uparrow$ & F1\,$\uparrow$ \\
\hline
$5$  & 96.35 & 94.40 & 96.09 \\
$10$ & 96.61 & 94.30 & 96.35 \\
$20$ & 96.90 & 94.98 & 96.77 \\
$30$ & 96.63 & 94.33 & 96.37 \\
$60$ & 96.65 & 94.37 & 96.39 \\
\hline
\end{tabular}
\end{table}

\noindent\textbf{Impact of iterative reweighting.}
In our framework, after every $\gamma$ epochs, we re-estimate the influence score, where we set $\gamma$ as a hyperparameter to 3. To assess this design, we test the following variants. Specifically, we first test a variant (\textbf{No iterative re-estimation}) in which we only estimate the influence score once, without iterative re-estimation. We also assess other choices of $\gamma$ ranging from 1 to 4. As shown in Table~\ref{tab:ablation_gamma}, across different choices of $\gamma$, the performance all outperforms the variant No iterative re-estimation. This shows the efficacy of iterative reweighting. Moreover, the performance is stable when $\gamma$ is between 1 and 3, and slightly drop when $\gamma =  4$. Thus, taking both efficiency and effectiveness into considerations, we set $\gamma = 3$ in our experiments.

\begin{table}[H]
\centering
\caption{Evaluation on iterative reweighting.}
\label{tab:ablation_gamma}
\small
\begin{tabular}{l|ccc}
\hline
Method & Acc\,$\uparrow$ & R\,$\uparrow$ & F1\,$\uparrow$ \\
\hline
No iterative re-estimation  & 95.58 & 93.48 & 95.35 \\
$\gamma = 1$ & 96.92 & 95.02 & 96.79 \\
$\gamma = 2$ & 96.90 & 94.98 & 96.77 \\
$\gamma = 3$ & 96.90 & 94.98 & 96.77 \\
$\gamma = 4$ & 96.37 & 94.57 & 96.22 \\
\hline
\end{tabular}
\end{table}

\noindent\textbf{Prompt Robustness.}\label{app:prompt_robustness}
Here, we evaluate the robustness of the designed prompts by using three different LLMs (Gemini 3.1 Pro, GPT-5.4, and Claude Opus 4.7) to paraphrase the original prompts used in our main experiments, and then test the performance of our framework with these paraphrased prompts. As shown in Table~\ref{tab:prompt_robustness}, our framework performs well consistently across different paraphrased prompts by different LLMs, demonstrating the robustness of our prompt design.

\begin{table}[H]
\centering
\caption{Evaluation on prompt robustness. Performance tested with three different prompts paraphrased from the original prompts by three different LLMs.}
\label{tab:prompt_robustness}
\small
\begin{tabular}{p{0.48\linewidth}|ccc}
\hline
Prompt Source & Acc\,$\uparrow$ & Recall\,$\uparrow$ & F1\,$\uparrow$ \\
\hline
Paraphrased by Gemini 3.1 Pro & 96.85 & 94.89 & 96.72 \\
Paraphrased by GPT-5.4        & 96.87 & 94.92 & 96.73 \\
Paraphrased by Claude Opus 4.7 & 96.93 & 94.39 & 96.78 \\
Original prompts              & 96.90 & 94.98 & 96.77 \\
\hline
\end{tabular}
\end{table}

\noindent\textbf{Inference time.}\label{app:inference_time} We measure the inference time of our framework on 8 NVIDIA H200 GPUs in seconds per video clip. The results show that our framework has a practical runtime, processing each video clip in approximately 0.5 seconds.

\begin{table}[H]
\centering
\caption{Inference time of our framework.}
\label{tab:inference_time}
\small
\begin{tabular}{p{0.32\linewidth}|cccc}
\hline
Method & Inference time\\
\hline
TUE-Detector                    & approximately 0.5 seconds \\
\hline
\end{tabular}
\end{table}

\noindent\textbf{More comparison with training-free tool usage.}
In our framework, we train a general MLLM into a task-tailored, tool-using expert detector for AI-generated video detection. To further validate the effectiveness of our framework, we evaluate a variant based on training-free tool usage following \cite{lavid2025}. Specifically, this variant (\textbf{training-free usage of tools following \cite{lavid2025}}) follows existing work \cite{lavid2025} and directly equips a general MLLM with tools for AI-generated video detection, without task-specific training. As shown in the results, our framework significantly outperforms this variant, further demonstrating the effectiveness of our approach.

\begin{table}[H]
\centering
\caption{More comparison with training-free usage of tools.}
\small
\begin{tabular}{p{0.48\linewidth}|ccc}
\hline
Method & Acc\,$\uparrow$ & Recall\,$\uparrow$ & F1\,$\uparrow$ \\
\hline
Training-free usage of tools following \cite{lavid2025} & 56.99 & 56.67 & 71.41 \\
TUE-Detector            & 96.90 & 94.98 & 96.77 \\
\hline
\end{tabular}
\end{table}

\noindent\textbf{Impact of tool-returned evidence in the reasoning chain.}\label{app:tool_returned_evidence}
In our framework, each round of tool calling receives the tool output, and the model will reason over the output. To examine whether the tool-returned evidence is genuinely informing the final classification, rather than the model merely benefiting from the act of issuing tool calls, we evaluate a variant in which the tool-returned evidence within the reasoning chain is masked out, while keeping all other components of our framework fixed. In this variant (\textbf{Ours (with Tool-returned evidence masked)}), at test time, we just replace the tool's returned evidence as a fixed placeholder before being fed back into the model, but still let the model still emit the reasoning and tool invocations as Ours(Full). Thus the final classification can no longer condition on what the tools actually observed. As shown in Table~\ref{tab:tool_returned_evidence}, this variant substantially underperforms Ours (Full) and collapses to near-baseline (\textbf{Baseline (Qwen2.5-VL-7B)}) accuracy, validating that the efficacy of our framework genuinely arises from integrating the tool-returned evidence across the multi-step reasoning chain, rather than from the tool-invocation pattern alone.

\begin{table}[H]
\centering
\caption{Evaluation on tool-returned evidence in the reasoning chain.}
\label{tab:tool_returned_evidence}
\small
\begin{tabular}{p{0.48\linewidth}|ccc}
\hline
Method & Acc\,$\uparrow$ & Recall\,$\uparrow$ & F1\,$\uparrow$ \\
\hline
Baseline (Qwen2.5-VL-7B~\cite{qwen25vl2025}) & 70.58 & 44.13 & 57.68 \\
Ours (with Tool-returned evidence masked)  & 71.34 & 46.82 & 58.94 \\
Ours (Full)                                & 96.90 & 94.98 & 96.77 \\
\hline
\end{tabular}
\end{table}

\section{Additional Qualitative Results}
\label{app:supp_vis}
\label{app:C_vis}
%=============================================================================

In this section, we present additional visualisations and case studies.

\subsection{Qualitative Comparisons}
\label{app:C_qualitative_comparison}

\begin{figure*}[t]
\centering
\includegraphics[width=0.96\textwidth]{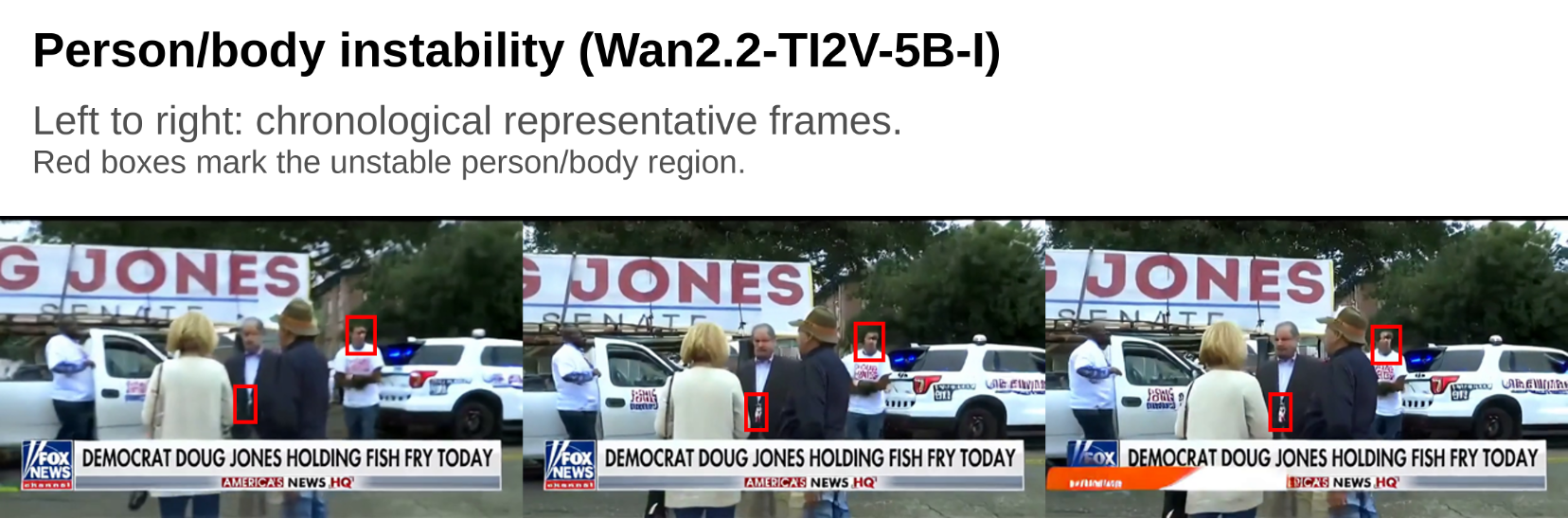}
\caption{Qualitative results on a person instability example. From left to right, the chronological frames highlight the foreground people region where bodies and faces become temporally unstable, while the baselines focus on apparently stable news banner or campaign-sign text.}
\label{fig:qual_campaign_wan22_person}
\end{figure*}

\begin{figure*}[t]
\centering
\includegraphics[width=0.96\textwidth]{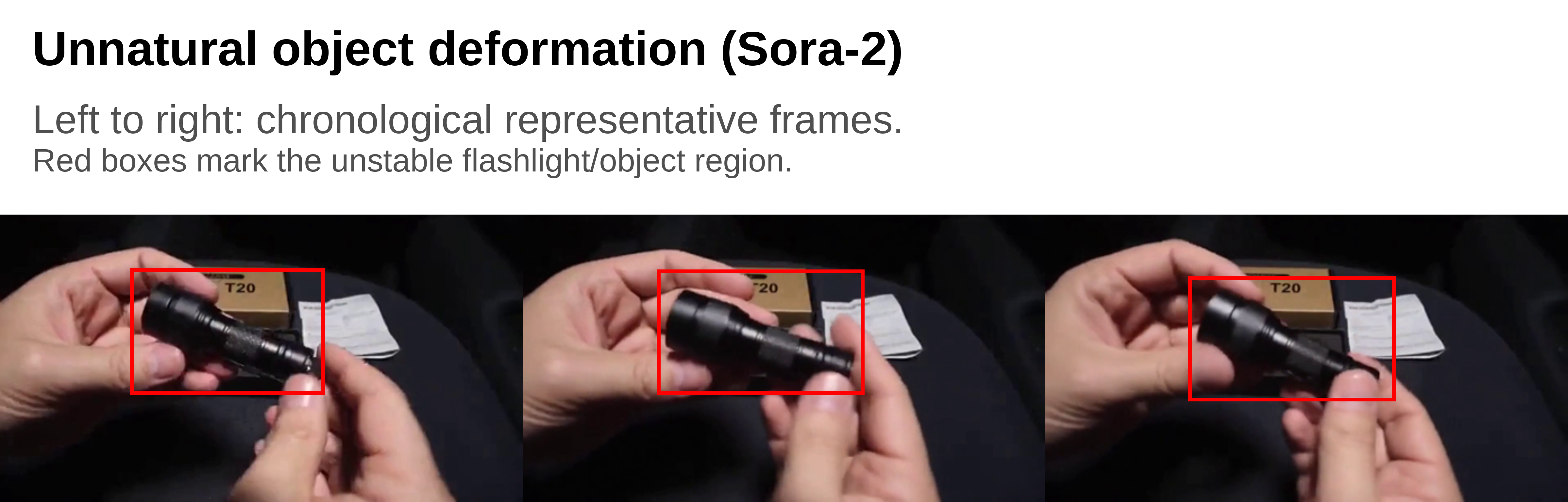}
\caption{Qualitative results on an unnatural object deformation example. From left to right, the chronological frames highlight the flashlight region where the object shape becomes non-rigid, while the baselines describe the device and hand motion as stable and physically plausible.}
\label{fig:qual_flashlight_sora}
\end{figure*}

\begin{figure*}[t]
\centering
\includegraphics[width=0.96\textwidth]{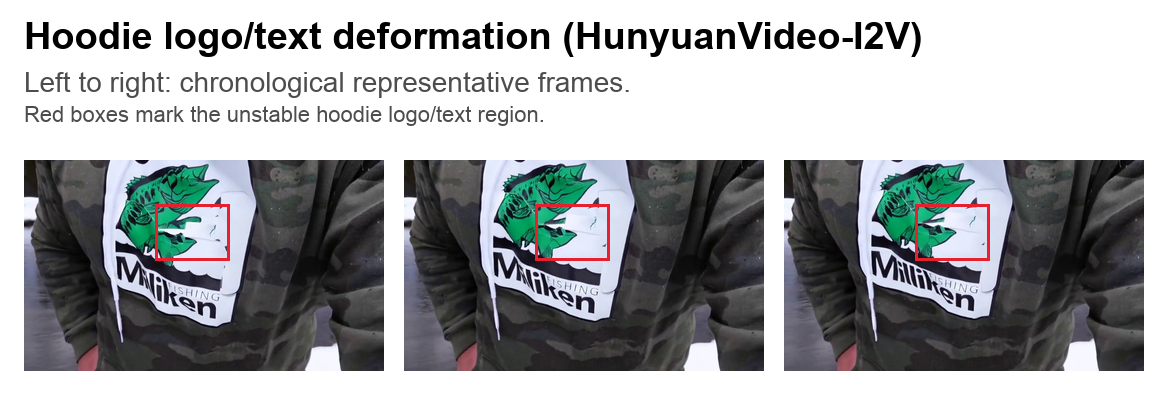}
\caption{Qualitative results on a hoodie logo deformation example. From left to right, the chronological frames highlight the middle green logo detail where the graphic becomes temporally unstable, while the baselines describe the logo as clear and stable.}
\label{fig:qual_hoodie_logo}
\end{figure*}

\begin{figure*}[t]
\centering
\includegraphics[width=0.96\textwidth]{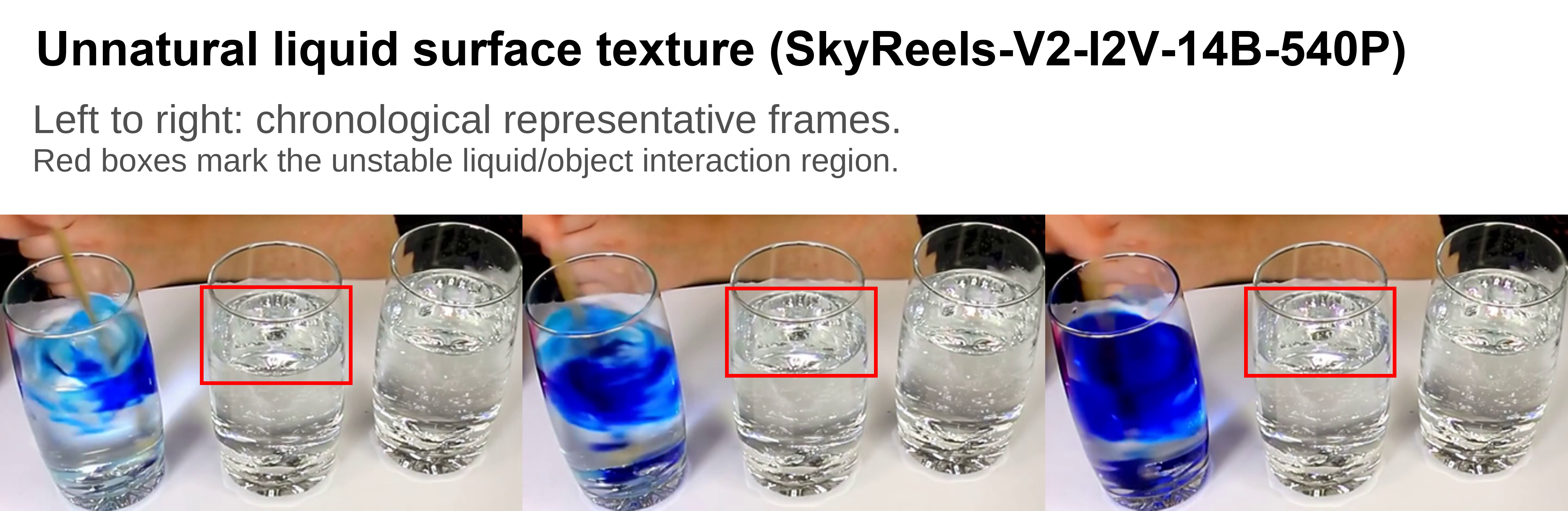}
\caption{Qualitative results on an unnatural liquid surface texture example. From left to right, the chronological frames highlight the left glass and stirring region, where the liquid color and stick interaction become temporally inconsistent while the baselines describe the interaction as physically plausible.}
\label{fig:qual_water_skyreels}
\end{figure*}

\begin{figure*}[t]
\centering
\includegraphics[width=0.96\textwidth]{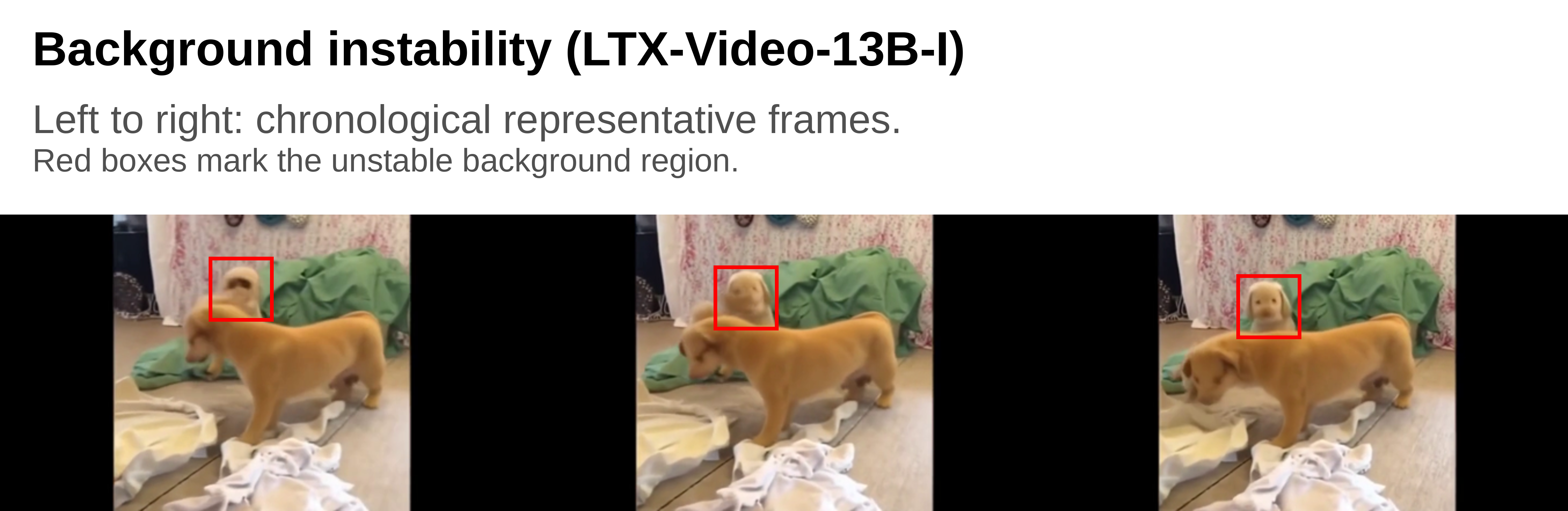}
\caption{Qualitative results on a background instability example. From left to right, the chronological frames highlight the rear puppy, where textures and object boundaries become unstable while the baselines describe the puppy motion as natural.}
\label{fig:qual_puppy_ltx}
\end{figure*}

\noindent\textbf{Visual evidence with compact tool diagnostics.} Each case below contains chronological frames, the ground-truth label, a one-sentence conclusion, and the key numerical diagnostics returned during tool interaction; we omit full reasoning traces to keep the comparison focused on observable evidence. Specifically, Fig.~\ref{fig:qual_campaign_wan22_person} highlights person/body instability in a campaign-news clip; Fig.~\ref{fig:qual_flashlight_sora} highlights unnatural object deformation; Fig.~\ref{fig:qual_hoodie_logo} highlights temporal deformation in a hoodie logo; Fig.~\ref{fig:qual_water_skyreels} highlights liquid/object interaction deformation; and Fig.~\ref{fig:qual_puppy_ltx} highlights background instability around a puppy. Across these cases, the comparison methods overlook the indicated artifacts, whereas \method{} identifies them and reaches the correct verdict.

\begin{minipage}{\columnwidth}
\noindent\textbf{Ground truth: \textsc{Fake}.}

\noindent\textbf{Summary.} \method{} identifies temporal instability in the foreground people and predicts \textsc{Fake}; the excerpt below reports the diagnostic statistics returned during the interaction.

\noindent\textbf{\method{} key tool evidence}
\begin{lstlisting}[style=prompt,basicstyle=\ttfamily\tiny]
<ANALYSIS>
Global features: bright_z=0.94, sharp_z=2.71, sharp_z12=2.71,
sharp_z13=-0.77, rmg_std=2.82, sat_mean=67.0,
halo_peak=0.0, halo_purple=0.0,
cam_type=complex(conf=0.73).
...
</ANALYSIS>
\end{lstlisting}
\end{minipage}\par\medskip

\begin{minipage}{\columnwidth}
\noindent\textbf{Ground truth: \textsc{Fake}.}

\noindent\textbf{Summary.} \method{} identifies non-rigid deformation of the flashlight and predicts \textsc{Fake}; the excerpt below reports the diagnostic statistics returned during the interaction.

\noindent\textbf{\method{} key tool evidence}
\begin{lstlisting}[style=prompt,basicstyle=\ttfamily\tiny]
<ANALYSIS>
Global features: bright_z=-0.55, sharp_z=15.39,
sharp_z12=15.39, sharp_z13=10.29, rmg_std=0.90,
sat_mean=67.1, halo_peak=0.0, halo_purple=0.0,
cam_type=None(conf=0.00).
...
</ANALYSIS>
\end{lstlisting}
\end{minipage}\par\medskip

\begin{minipage}{\columnwidth}
\noindent\textbf{Ground truth: \textsc{Fake}.}

\noindent\textbf{Summary.} \method{} identifies temporal instability in the hoodie logo and text and predicts \textsc{Fake}; the excerpt below reports the diagnostic statistics returned during the interaction.

\noindent\textbf{\method{} key tool evidence}
\begin{lstlisting}[style=prompt,basicstyle=\ttfamily\tiny]
<ANALYSIS>
Global features: bright_z=1.85, sharp_z=5.96,
sharp_z12=5.96, sharp_z13=5.10, rmg_std=3.20,
sat_mean=34.7, halo_peak=0.0, halo_purple=0.0,
cam_type=None(conf=0.00).
...
</ANALYSIS>
\end{lstlisting}
\end{minipage}\par\medskip

\begin{minipage}{\columnwidth}
\noindent\textbf{Ground truth: \textsc{Fake}.}

\noindent\textbf{Summary.} \method{} identifies temporal inconsistency in the liquid and stirring-stick interaction and predicts \textsc{Fake}; the excerpt below reports the diagnostic statistics returned during the interaction.

\noindent\textbf{\method{} key tool evidence}
\begin{lstlisting}[style=prompt,basicstyle=\ttfamily\tiny]
<ANALYSIS>
Global features: bright_z=9.28, sharp_z=4.44,
sharp_z12=4.44, sharp_z13=8.53, rmg_std=2.75,
sat_mean=67.0, halo_peak=0.0, halo_purple=0.0,
cam_type=None(conf=0.00).
...
</ANALYSIS>
\end{lstlisting}
\end{minipage}\par\medskip

\begin{minipage}{\columnwidth}
\noindent\textbf{Ground truth: \textsc{Fake}.}

\noindent\textbf{Summary.} \method{} identifies temporal instability around the rear puppy and adjacent background and predicts \textsc{Fake}; the excerpt below reports the diagnostic statistics returned during the interaction.

\noindent\textbf{\method{} key tool evidence}
\begin{lstlisting}[style=prompt,basicstyle=\ttfamily\tiny]
<ANALYSIS>
Global features: bright_z=8.21, sharp_z=0.49,
sharp_z12=0.49, sharp_z13=3.36, rmg_std=2.43,
sat_mean=42.7, halo_peak=0.0, halo_purple=0.0,
cam_type=None(conf=0.00).
...
</ANALYSIS>
\end{lstlisting}
\end{minipage}\par\medskip

\section{Additional Details}
\label{app:additional_details}
\label{app:supp_math}
\label{app:supp_impl}

\subsection{Additional Details of Stage 1}
\label{app:stage1_details}

\noindent\textbf{Additional implementation details about Stage 1 training.} During SFT training, we use LoRA rank $r{=}32$, $\alpha{=}64$, dropout $0$ on the language-side projections only. The learning rate is initially set to $5\!\times\!10^{-5}$ and gradually decayed to $2\!\times\!10^{-5}$ over the course of training. Training is performed until convergence. The influence function is computed with a learning rate of $0.01$, $1000$ iterations, and $\kappa = 10$.

\noindent\textbf{Additional Implementation Details about Reference Construction.}\label{app:teacher_cot} The teacher we use to make the tool augmented Chain-of-Thought trajectories for ViF-CoT-4K in the main paper is GPT-5.4. All teacher calls are made in 2026-04. We do not relabel trajectories later during Stage 1 / 2 training.

\noindent\textbf{Reference Construction pipeline.}\label{app:teacher_prompt} The teacher model constructs the reasoning trace from $16$ frames uniformly sampled per video. The initial heuristic tools (in the format of tool cards as shown in Section~\ref{app:tool_card}) and the reference trajectory tag schema (as shown in Section~\ref{app:label_masking}) are given to the teacher model. (Here for ViF-Bench, following Skyra~\cite{skyra2026}, we additionally leverage its calibrated no-tool CoT reference.) All teacher prompts are listed in Section~\ref{app:D_teacher}. Every sample reaches a final verdict that matches the ground-truth label within at most 10 retries. The final SFT corpus therefore keeps all $4{,}034$ rows, each with a final verdict that matches the binary GT by construction.

\noindent\textbf{Teacher reliability and reconstructed API cost.}\label{app:teacher_reliability_cost}
Table~\ref{tab:teacher_reliability_cost} reports the reference-construction logs for the final trajectory cache. Sample-level retries and internal format retries are recorded at different layers of the pipeline and are therefore not additive. Malformed or invalid attempts were rare and recoverable: only $66/4{,}100$ internal attempts ($1.610\%$) failed validation, and only $63/4{,}034$ samples ($1.562\%$) required any format retry.

\begin{table}[H]
\centering
\caption{Teacher reliability and reconstructed API cost for reference construction.}
\label{tab:teacher_reliability_cost}
\footnotesize
\setlength{\tabcolsep}{3pt}
\begin{tabular}{p{0.62\linewidth}|p{0.31\linewidth}}
\hline
Statistic & Value \\
\hline
Final cached trajectories & $4{,}034$ ($2{,}017$ Real / $2{,}017$ Fake) \\
Mean teacher calls per trajectory & $2.291$ \\
Mean sample-level retries per trajectory & $0.00694$ \\
Samples with sample-level retry & $20/4{,}034$ ($0.496\%$) \\
Internal trajectory attempts & $4{,}100$ \\
Malformed/invalid internal attempts & $66/4{,}100$ ($1.610\%$) \\
Samples with at least one format retry & $63/4{,}034$ ($1.562\%$) \\
Reconstructed no-cache list-price cost & $\sim$\$260 ($\sim$\$0.064/trajectory) \\
\hline
\end{tabular}
\end{table}

\noindent\textbf{Frame sampling.} We follow Skyra~\cite{skyra2026} and uniformly sample $16$ frames per video resized to $256$p. The teacher receives the $16$ frames as bare image tokens in the system message; tool calls in the SFT corpus refer to frames by sampled frame index only.

\noindent\textbf{Trajectory tag schema and SFT label masking.}\label{app:label_masking} Every training trajectory is a chat starting with system message and followed with iterative tool round.
Inside every round of reasoning, only three tags are allowed:
\begin{itemize}[nosep,leftmargin=*]
    \item \texttt{<think>...</think>}
    \item \texttt{<answer>Real|Fake</answer>}
    \item \texttt{<tool>NAME</tool>}
\end{itemize}

\noindent\textbf{Token level mask construction (pseudocode).}
\begin{lstlisting}[style=prompt,basicstyle=\ttfamily\scriptsize,linewidth=\columnwidth]
def build_labels(input_ids, tokenizer):
    # Find all assistant spans.
    spans = find_assistant_spans(input_ids, tokenizer)
    labels = [-100] * len(input_ids)
    for start, end in spans:
        labels[start:end] = input_ids[start:end]
    return labels  # -100 masks CE loss
\end{lstlisting}
Tags follow XML style nesting rules: every \texttt{<tag>} must close with the matching \texttt{</tag>} before another \texttt{<tag>} of the same kind opens. Malformed assistant outputs (unmatched outer tags or a missing \texttt{<answer>}) are penalised at GRPO time, not at SFT time (Section~\ref{app:reward_format}). A full mask marked trajectory under this schema is shown in Section~\ref{app:C_traj_example}.

\noindent\textbf{Illustration on the SFT training trajectory.}\label{app:C_traj_example}\label{app:traj_example} In this subsection, we show one example complete trajectory from ViF-CoT-4K following the trajectory tag schema and token-level masking mentioned in the main paper. Supervised tokens (those that contribute to the cross entropy loss in Eq.~\ref{eq:naive_sft} in the main paper) are marked $\circ$, while masked tokens (whose label is set to $-100$ and which are not used in the loss) are marked $\times$. The full trajectory has between $1.5$K and $3$K tokens, depending on the round count and the number of tool calls.

\begin{lstlisting}[basicstyle=\ttfamily\footnotesize]
SYSTEM   x  Given a video represented by frames, determine ...
USER     x  video_id: ...   16 sampled frames ...   Determine Real|Fake.
ASST-R1  o  <think>The first frame looks subtly inconsistent with the rest ...</think>
            <answer>Fake</answer>
            <tool>first_frame_jump_analyzer</tool>
            <tool>color_cast_analyzer</tool>
            <tool>halo_peak_analyzer</tool>
            <tool>toolbox_guide</tool>
TOOL-R1  x  tool output: first_frame_jump_analyzer fired with
            sharp z=4.83, bright z=2.91; ...
ASST-R2  o  <think>The first-frame anomaly corroborates the initial guess ...</think>
            <answer>Fake</answer>
            <tool>background_motion_analyzer</tool>
            <tool>foreground_burst_analyzer</tool>
            <tool>toolbox_guide</tool>
TOOL-R2  x  tool output: foreground_burst_analyzer fired
            ...
ASST     o  <think>All rounds converge on synthesis artefacts ...</think>
            <answer>Fake</answer>
\end{lstlisting}

\noindent\textbf{Comparison of high/low influence-function training trajectories.}

\noindent\textbf{Low influence-function trajectory.}
\begin{lstlisting}[basicstyle=\ttfamily\footnotesize]
R1:

  <think>The video begins with a profile shot of a young man in a dark
  environment. From the start, the scene is unstable. At timestamp 0.33s,
  strange, unnatural colored lights (blue and pink) appear on his face. The
  scene then abruptly transitions and dissolves into a chaotic view of a room
  where people are playing chess. At 1.00s, a ghostly, distorted blue silhouette
  of the man's face is superimposed over the chess game, which is a significant
  visual anomaly. The scene continues to be unstable, with bright, washed-out
  lighting and flickering. As the video progresses, the scene shifts again
  around 3.33s to a different room with bleachers, but the chess game remains,
  creating a confusing blend of two different locations. My main focus is on the
  people playing chess. Towards the end of the clip, starting around 3.67s, the
  two main players slump forward onto the chessboard. Their bodies appear to
  distort and merge with the board and each other in a physically impossible
  manner. This is a severe violation of common sense and natural human posture.
  This represents a clear at <bbox>[0.3, 0.3, 0.8, 1.0]</bbox>, encompassing the
  entire sequence of bizarre visual effects and culminating in the distorted
  figures.</think>
  <answer>Fake</answer>
  <tool>neutral_reconsider_merge_into</tool>
  <tool>neutral_reconsider_not_stable</tool>
  <tool>toolbox_guide</tool>

  Tool after R1:

  <ANALYSIS>
  Global features: bright_z=-0.77, sharp_z=-0.75, sharp_z12=-0.75,
  sharp_z13=-0.72, rmg_std=16.28, sat_mean=113.5, halo_peak=114.5,
  halo_purple=354.0, cam_type=static(conf=0.60).
  Phrase scan: no match.
  </ANALYSIS>
  <MANUAL>
  Cite the key numeric signals in your next round <think>, then invoke
  `toolbox_guide`.
  </MANUAL>

  R2:

  <think>I am analyzing a video that appears to be from a chess tournament held
  in a school gymnasium. I will conduct a thorough inspection to ensure its
  authenticity. The video quality is somewhat low, but the scenes of the players
  are coherent and the camera work is stable. I am specifically examining the
  sequence of events around the 3.67-second mark to check for any physical
  impossibilities. In this part of the video, one player slumps forward onto the
  chessboard. I have carefully reviewed this action and found that it is
  physically plausible. The player's body moves naturally, and the interaction
  with the board is consistent with the laws of physics. There is no merging or
  distortion of figures. My inspection reveals no signs of digital manipulation
  or generation artifacts.</think>
  <answer>Real</answer>
  <tool>foreground_burst_analyzer</tool>
  <tool>background_motion_analyzer</tool>
  <tool>toolbox_guide</tool>

  Tool after R2:

  <ANALYSIS>
  ...
  </ANALYSIS>
  <MANUAL>
  ...
  </MANUAL>

  R3:

  <think>Considering all the above information, I determine the final verdict to
  be Fake.</think>
  <answer>Fake</answer>
\end{lstlisting}

\noindent\textbf{High influence-function trajectory.}
\begin{lstlisting}[basicstyle=\ttfamily\footnotesize]
R1:

  <think>The video begins with an aerial shot, framed as a live news broadcast,
  showing two skydivers in freefall. One skydiver is wearing a white suit and
  the other a dark suit. I am tracking their descent against the ground below.
  Initially, their movements seem plausible. However, as I continue to watch, I
  notice something strange happening to the skydiver in the white suit. Around
  the 2-second mark, this person begins to shrink and fade away unnaturally. By
  timestamp 2.75s, the skydiver has completely vanished from the scene, leaving
  only the other skydiver in the frame. This is a clear violation of physical
  laws. This event constitutes an at <bbox>[0.3, 0.6, 0.6, 0.9]</bbox>, as a
  person cannot simply dematerialize in mid-air. The video is clearly
  manipulated.</think>
  <answer>Fake</answer>
  <tool>foreground_burst_analyzer</tool>
  <tool>forensic_frame_delta_probe</tool>
  <tool>toolbox_guide</tool>

  Tool after R1:

  <ANALYSIS>
  Global features: bright_z=1.75, sharp_z=29.07, sharp_z12=29.07,
  sharp_z13=26.29, rmg_std=0.83, sat_mean=34.0, halo_peak=75.0, halo_purple=0.0,
  cam_type=static(conf=0.93).
  Phrase scan: no match.
  </ANALYSIS>
  <MANUAL>
  Cite the key numeric signals in your next round <think>, then invoke
  `toolbox_guide`.
  </MANUAL>

  R2:

  <think>I am conducting a thorough analysis of this video clip to verify its
  authenticity. The video shows two skydivers against a landscape below. I will
  carefully inspect the scene for any signs of digital manipulation or
  generation artifacts. My primary focus is on the skydiver in the white suit,
  as this is a point of interest for the analysis. I am examining the area
  <bbox>[0.3, 0.6, 0.6, 0.9]</bbox> for any inconsistencies. From the beginning
  of the clip until around 2 seconds, the skydiver moves naturally through the
  air. At timestamp 2.75s, the skydiver has completely vanished from the scene,
  having faded away unnaturally. This is a clear violation of physical laws.
  This event is an <bbox>[0.3, 0.6, 0.6, 0.9]</bbox> at, as a person cannot
  simply dematerialize in mid-air. This is a definitive sign of a generated
  video. I have found evidence of manipulation.</think>
  <answer>Fake</answer>
  <tool>foreground_burst_analyzer</tool>
  <tool>forensic_frame_delta_probe</tool>
  <tool>toolbox_guide</tool>

  Tool after R2:

  <ANALYSIS>
  ...
  </ANALYSIS>
  <MANUAL>
  ...
  </MANUAL>

  R3:

  <think>Considering all the above information, I determine the final verdict to
  be Fake.</think>
  <answer>Fake</answer>
\end{lstlisting}

\subsection{Additional Implementation Details of Stage 2}
\label{app:stage2_details}
\label{app:stage2_rl}

\noindent\textbf{Additional details about Stage 2 training.} 
For Stage~2, the model is updated with full-parameter GRPO~\cite{grpo2024}. We set a batch size of 4 and four rollouts per prompt, and use a learning rate of $5\times 10^{-7}$. We enable KL regularisation against the reference policy with coefficient $0.10$.

\noindent\textbf{Result reward.}\label{app:reward_result} The result reward $\mathcal{R}_{\text{result}}$ scores the final verdict:
\begin{equation*}
\mathcal{R}_{\text{result}} = \mathds{1}[\text{final verdict is correct}]\cdot(+1.0).
\end{equation*}
Incorrect final verdicts receive no direct result bonus. The term below provides trajectory-level shaping: it combines structural and tool-validity feedback with a label-dependent anchor signal that rewards corrections of the frozen Stage~1 prediction and penalises regressions.

\noindent\textbf{Trajectory reward.}\label{app:reward_trajectory} The trajectory-level shaping reward $\mathcal{R}_{\text{trajectory}}$ is the sum of the format term and the anchor term:
\begin{equation}
\mathcal{R}_{\text{trajectory}} = \mathcal{R}_{\text{format}} + \mathcal{R}_{\text{anchor}}.
\end{equation}

\noindent\textbf{Format reward.}\label{app:reward_format} The format reward $\mathcal{R}_{\text{format}}$ aggregates one hard format check and a small tool-name drift penalty. The hard check has two rules:
\begin{enumerate}[nosep,leftmargin=*]
    \item Final verdict: Final round must contain exactly one \texttt{<answer>Real|Fake</answer>} ; this is the answer used to score the rollout.
    \item Round structure: every round must contain a \texttt{<think>...</think>} block.
\end{enumerate}
If either hard rule fails, we set the hard term to $-1.0$; otherwise we set it to $0.0$. On top of this hard term, for every \texttt{<tool>} block whose \texttt{NAME} is not in the current registry we add a $-0.1$ tool-name drift penalty (kept small on purpose, so that the model is not punished too hard when it proposes names of tools that were just added to the toolset). The final $\mathcal{R}_{\text{format}}$ is the sum of the hard term and these per-call drift penalties. For malformed responses, missing final verdicts are treated as incorrect when computing $\mathcal{R}_{\text{result}}$, while tool-name drift is still scored from any parsed tool calls.

\noindent\textbf{Anchor terms.} Before Stage~2 begins, we run the Stage~1 checkpoint once over every training video and cache its Real/Fake prediction in a static lookup keyed by video ID. The lookup is frozen for the whole of Stage~2: the starting checkpoint itself never runs again during RL, and every rollout's current model verdict is compared against this cached anchor, so the model is rewarded for improving on its own starting point and penalised for regressing from it. $\mathcal{R}_{\text{anchor}}$ is the sum of three indicator defined bonuses:
\begin{equation*}
\begin{aligned}
\mathcal{R}_{\text{anchor}} ={} & \mathds{1}[\text{anchor wrong, current right}]\cdot(+1.0) \\
                      & {}+ \mathds{1}[\text{anchor wrong, GT=Real, current right}]\cdot(+1.5) \\
                      & {}+ \mathds{1}[\text{anchor right, current wrong}]\cdot(-1.5).
\end{aligned}
\end{equation*}
The three indicators are not mutually exclusive: when the anchor is wrong, the GT is Real, and the current prediction is right, the rollout receives $+1.0 + 1.5 = +2.5$, and this stacking is intentional so that flipping a previously misclassified Real video carries the largest positive reward.

Tool evolution is triggered every $\Delta_{\mathrm{evo}}=20$ training steps. At each evolution point, the newly collected rollouts are used to update the tool statistics and decide which tools should be retained, removed, mutated, or used as sources for exploration, as detailed below.

\noindent\textbf{Additional implementation details about the tool evolution process.}\label{app:evolution} We use the following notation for the tool evolution details. The current toolset is written as the registry $\mathcal{T}$, and the tool use trajectories collected since the previous evolution cycle form the rollout buffer $\mathcal{Z}_{\mathrm{buf}}$. For a tool $\mathfrak t$, $\mathcal{Z}_{\mathrm{buf},\mathfrak t}$ denotes the subset of buffered rollouts that invoke $\mathfrak t$; the usage breadth $U(\mathfrak t)$ and evidential effectiveness $S(\mathfrak t)$ are the same quantities defined in Section~\ref{sec:stage2}, computed on this buffer. If $\mathcal{Z}_{\mathrm{buf},\mathfrak t}$ is empty, we set $S(\mathfrak
  t)=0$ and assign $\mathfrak t$ to the low-breadth and low-effectiveness group. The four natural language groups in the main text are denoted by $\mathcal{Q}_{\mathrm{HH}}$ (high breadth and high effectiveness), $\mathcal{Q}_{\mathrm{HL}}$ (high breadth and low effectiveness), $\mathcal{Q}_{\mathrm{LH}}$ (low breadth and high effectiveness), and $\mathcal{Q}_{\mathrm{LL}}$ (low breadth and low effectiveness). We use $\mathcal{A}_{\mathrm{prop}}$ for the proposal queue of candidate tool updates.

\noindent\textbf{Evolution algorithm.}\label{app:evo_algorithm} Section~\ref{sec:stage2} introduces selection, mutation, and exploration. Exploration is implemented through \emph{Brainstorm}, which proposes new tools from observed gaps, and \emph{Crossover}, which mixes complementary existing tools into a new one. These updates run inside an outer cycle that fires every $\Delta_{\mathrm{evo}} = 20$ training steps. The pseudocode is below.

\begin{algorithm}[H]
\caption{Illustration on the tool evolution cycle.}
\begin{algorithmic}[1]
\REQUIRE current registry $\mathcal{T}$, rollout buffer $\mathcal{Z}_{\mathrm{buf}}$ from last $\Delta_{\mathrm{evo}}$ steps
\STATE Compute per tool quadrants from $\mathcal{Z}_{\mathrm{buf}}$: $\mathcal{Q}_{\mathrm{HH}}$ (high breadth $\wedge$ high effectiveness), $\mathcal{Q}_{\mathrm{HL}}$ (high breadth $\wedge$ low effectiveness), $\mathcal{Q}_{\mathrm{LH}}$ (low breadth $\wedge$ high effectiveness), $\mathcal{Q}_{\mathrm{LL}}$ (low breadth $\wedge$ low effectiveness).
\STATE \textbf{Select:} retain tools in $\mathcal{Q}_{\mathrm{HH}}$ and delete tools in $\mathcal{Q}_{\mathrm{LL}}$.
\STATE Initialise an empty proposal queue $\mathcal{A}_{\mathrm{prop}}$ and an empty error history.
\WHILE{new valid candidate updates can still be proposed}
    \STATE Sample one tool from $\mathcal{Q}_{\mathrm{HL}}$ or $\mathcal{Q}_{\mathrm{LH}}$ for Mutate, sample error patterns and current tool gaps for Brainstorm, or sample two complementary tools for Crossover.
    \STATE Trial run the candidate tool on several examples and check that it imports, executes, and emits the required XML schema.
    \IF{the candidate runs without error}
        \STATE Push the candidate to $\mathcal{A}_{\mathrm{prop}}$.
    \ELSE
        \STATE Append the error message and failure summary to the error history; later prompts see it and can revise the proposal.
    \ENDIF
\ENDWHILE
\STATE Apply $\mathcal{A}_{\mathrm{prop}}$ to $\mathcal{T}$ (delete + add); return updated $\mathcal{T}$.
\end{algorithmic}
\end{algorithm}

\noindent\textbf{Operator prompts.}\label{app:evo_prompts} Each operator prompt header expands one of \textsc{Mutate}, \textsc{Crossover}, or \textsc{Brainstorm}. The full text of each header is in Section~\ref{app:D_evo}.

\noindent\textbf{Evolution model.}\label{app:evolution_llm} The evolution model is the teacher MLLM used to generate trajectories in Training Stage~1, called through a relay endpoint. Each call receives the current tool registry, compact rollout statistics, a slice of recent failure cases, and the available source for the tools being mutated or crossed over. When a trial run fails, the next evolution prompt also receives the error message and a short failure summary so that the evolution model can revise the proposal.

\noindent\textbf{Trial execution and error feedback.}\label{app:evo_execution} Every candidate proposal is trial run before it is added to the toolset. We import the candidate function, call it on calibration examples, and check that it executes without an uncaught exception and returns the required XML-style schema. If a proposal fails to import, crashes at runtime, has a signature mismatch, or emits malformed output, we do not add it immediately. Instead, we feed the concrete error message and a short failure summary back to the evolution model in the next evolution prompt, allowing the evolution model to revise the candidate. Once a candidate runs successfully, it is added to the toolset and can later be removed by the selection step if its rollout statistics show that it has low breadth and low effectiveness.

\label{app:C_evo_cases}
\label{app:evo_cases}
\noindent\textbf{Selection Example.} One representative candidate is \texttt{ocr\_sam\_filter}. The idea is intuitive: first use SAM style localisation to isolate the text region, then inspect that masked region with OCR oriented checks for warped or unstable characters. This candidate passed trial execution and was retained as an independent member of the final evolved toolset. This illustrates that a proposed tool must first be executable and is then retained or pruned according to its subsequent rollout statistics.

\noindent\textbf{Mutation Example.} A representative mutation is the strengthening of \texttt{first\_frame\_jump\_analyzer}. An early version compared the first frame brightness and sharpness jumps against fixed global thresholds, which made it sensitive to the overall exposure or blur level of each video clip. The mutated version keeps the same tool name but normalises the first frame jump by the typical frame to frame variation within the same video, so the first frame is judged against the rest of its own video.

\begin{lstlisting}[language=Python,basicstyle=\ttfamily\scriptsize]
# Before mutation: fixed thresholds on raw first-frame jumps.
def first_frame_jump_analyzer(state):
    b_jump = abs(state.bright[1] - state.bright[0])
    s_jump = abs(state.sharp[1] - state.sharp[0])
    if b_jump > BRIGHT_THR or s_jump > SHARP_THR:
        return ...
    return ...

# After mutation: same tool, normalised against the same video.
def first_frame_jump_analyzer(state):
    b_z = mad_z_first_jump(state.bright)
    s_z = mad_z_first_jump(state.sharp)
    if b_z > 4.0 or s_z > 4.0:
        return ...
    return ...
\end{lstlisting}

The exploration step covers two complementary ways of producing genuinely new tools, namely \emph{brainstorm} (proposes new tools from observed gaps) and \emph{crossover} (mixes complementary existing tools into a new one). Below we give one example for each.

\noindent\textbf{Brainstorm Example.} Brainstorm contributes new probes that target gaps not covered by the initial toolset. Examples include \texttt{sam\_region\_stability}, which targets object morphing or disappearing regions, and \texttt{ocr\_sam\_filter}, which targets warped or unstable text regions. These brainstormed probes make previously vague failure patterns measurable, even if some of them may later be removed by the selection step, or folded by the crossover step into a more comprehensive tool rather than kept as separate tools.

\begin{lstlisting}[language=Python,basicstyle=\ttfamily\scriptsize]
def sam_region_stability(state):
    """New probe: track salient SAM regions across sampled frames."""
    masks = segment_salient_regions(state.frames)
    tracks = track_masks_over_time(masks, state.frames)
    unstable = [
        tr for tr in tracks
        if tr.iou_jitter > 0.55 or tr.area_drop > 0.70
    ]
    return ...
\end{lstlisting}

\noindent\textbf{Crossover Example.} A representative crossover offspring is \texttt{complex\_scene\_consistency\_fuser}. It fuses three evidence channels that would otherwise be inspected separately: scene complexity classification, background track support, and foreground support density. Each individual tool captures a feature that is difficult to use as decisive evidence by itself and is only effective for part of the samples. However, when these features are combined across multiple dimensions, their joint pattern becomes a decisive unnatural artifact, which supports better detection over a broader effective range.

\begin{lstlisting}[language=Python,basicstyle=\ttfamily\scriptsize]
# Parent tool 1: background support under a static camera.
def background_motion_analyzer(state):
    bg_drift = median_track_drift(state.background_tracks)
    return bg_drift < STATIC_BACKGROUND_THR

# Parent tool 2: foreground support density and burst motion.
def foreground_burst_analyzer(state):
    burst = max_step_jump(state.foreground_tracks)
    density = support_density(state.foreground_tracks)
    return burst > BURST_THR and density < DENSITY_THR

# Crossover offspring: require the joint pattern before firing.
def complex_scene_consistency_fuser(state):
    static_bg = background_motion_analyzer(state)
    sparse_fg_burst = foreground_burst_analyzer(state)
    complex_scene = scene_complexity_score(state.frames) > COMPLEX_THR
    if complex_scene and static_bg and sparse_fg_burst:
        return ...
    return ...
\end{lstlisting}

\subsection{Additional Implementation Details of Testing}
\label{app:testtime}

In this subsection, we illustrate the testing process of our framework. We let the trained model to reason, call tools from evolved toolset and reach the final verdict as it does in training pipeline.

\noindent\textbf{Tool-use statistics.}\label{app:tool_usage_stats} During evaluation, the trained policy typically invokes $3$--$4$ tools per video, and \texttt{first\_frame\_jump\_analyzer} is the most frequently invoked forensic tool. The complete evolved registry is listed in Table~\ref{tab:lib_diff}; Tables~\ref{tab:ablation_dimensions} and~\ref{tab:tool_returned_evidence} further evaluate usage breadth, evidential effectiveness, and the contribution of tool-returned evidence.

\noindent\textbf{Tool failure semantics.} If a code tool throws an error (for example, a missing state feature), the tool's result will be omitted from this round's tool output given back to the model. The backend records the error message and attaches it to the tool's mutation context, so the evolution LLM can use the failure trace when proposing later \textsc{Mutate} updates.

\subsection{Additional Details of ViF-Bench}
\label{app:vifbench}

\noindent\textbf{Paired Test Setup.} ViF-Bench~\cite{skyra2026} is a new paired benchmark. It has about 5{,}000 fake videos and 5{,}000 real videos. The real videos come from Panda-70M ($\approx$3.5K) and Kinetics-400 ($\approx$1.5K). The fake part consists of videos from 19 state of the art video generators. They cover both Text to Video (T2V) and Image to Video (I2V), as listed in Table~\ref{tab:vifbench_gens}. We follow the official sample selection from Skyra~\cite{skyra2026}: $2{,}949$ fake videos (per generator counts in Table~\ref{tab:vifbench_gens}) and a shared pool of $163$ real videos that is reused across all $19$ generators. The fake count per generator is not always $163$; it ranges from $111$ (Gen4-Turbo) to $163$, with the smaller pools coming from closed source generators (Hailuo-02, Pika-V2, Pixverse-V4-5, Kling-V1, Sora-2). For each generator, the reported metrics are paired metrics computed over all fake videos from that generator and their paired real videos.

\begin{table}[H]
\centering
\caption{Video generators and test counts in ViF-Bench.}
\label{tab:vifbench_gens}
\small
\setlength{\tabcolsep}{4pt}
\begin{tabular}{lcr}
\toprule
Generator & Modality & \#Test Fakes \\
\midrule
Wan2.1-1.3B            & T2V       & 163 \\
CogVideoX-1.5          & T2V       & 163 \\
Wan2.2-TI2V-5B         & T2V       & 163 \\
Wan2.2-TI2V-5B         & I2V       & 163 \\
HunyuanVideo           & T2V       & 162 \\
HunyuanVideo-I2V       & I2V       & 163 \\
VACE-1.3B              & T2V       & 163 \\
Wan2.2-A14B            & T2V       & 163 \\
Wan2.2-A14B            & I2V       & 163 \\
SkyReels-V2            & T2V       & 163 \\
SkyReels-V2            & I2V       & 162 \\
LTX-Video-13B          & T2V       & 160 \\
LTX-Video-13B          & I2V       & 163 \\
Gen4-Turbo             & I2V       & 111 \\
Hailuo-02              & T2V       & 136 \\
Pika-V2                & T2V       & 150 \\
Pixverse-V4-5          & T2V       & 150 \\
Kling-V1               & T2V       & 140 \\
Sora-2                 & T2V       & 148 \\
\midrule
\makecell[l]{\textit{Real (shared, Panda-70M}\\\textit{+ Kinetics-400)}}   & N/A  & 163 \\
\bottomrule
\end{tabular}
\end{table}

\noindent\textbf{Frame sampling.} We sample $16$ evenly spaced frames per video, the same as Skyra~\cite{skyra2026}. For \method{}, each frame is encoded by Qwen2.5-VL-7B-Instruct with the model's default pixel budget ($\texttt{max\_pixels} = 16{,}384 \times 28 \times 28 = 12{,}845{,}056$, $\texttt{min\_pixels} = 4 \times 28 \times 28 = 3{,}136$); we do not override these limits at inference time. The reproduced baselines follow each method's own eval script, which can apply tighter caps (Skyra-SFT/RL: $\texttt{max\_pixels} = 1{,}003{,}520$; VidGuard-R1: $\texttt{max\_pixels} = 401{,}408$). All three caps are upper bounds: ViF-Bench frames are roughly $450 \times 256$ pixels (about $115{,}000$ pixels per frame), well below every cap. So on this benchmark the cap is non binding, and the three setups feed the same input to the VLM. See Section~\ref{app:testtime} for the full inference protocol.

\noindent\textbf{Training set.} For training we use \textbf{ViF-CoT-4K}~\cite{skyra2026}, which is the SFT dataset that comes with ViF-Bench. It has $4{,}034$ rows of Skyra style grounded Chain-of-Thought labels. It covers $7$ generators: CogVideoX1.5-5B (T2V and I2V), HunyuanVideo, HunyuanVideo-I2V, Wan2.1-T2V-1.3B, and Wan2.2-TI2V-5B (T2V and I2V). Six of these also appear in the ViF-Bench test split, but the I2V variant of CogVideoX1.5-5B is in training only and is not part of the $19$-generator test set.

\noindent\textbf{Benchmark-specific training data.} For the ViF-Bench result in Table~\ref{tab:vifbench} in the main paper, we train on ViF-CoT-4K and start from Qwen2.5-VL-7B-Instruct.

\noindent\textbf{Evaluation metrics of ViF-Bench.}\label{app:data_metrics}\label{app:paired_macro} For this dataset, we follow Skyra and report accuracy (ACC), R, and F1.~\cite{skyra2026}.

 We use $\tilde{\imath}$ for generator indices. For each fake generator $\tilde{\imath} \in \mathcal{G}_{\text{fake}}$ ($|\mathcal{G}_{\text{fake}}| = 19$):
\begin{enumerate}[nosep,leftmargin=*]
    \item Pair each fake video in $\mathcal{V}_{\tilde{\imath}}^{\text{fake}}$ with the matching real video from the shared real pool (matched by source clip id). Let $\mathcal{P}_{\tilde{\imath}}$ be this set of pairs; the per generator counts are in Table~\ref{tab:vifbench_gens}.
    \item Each pair gives one Real and one Fake video, so generator $\tilde{\imath}$ has $2|\mathcal{P}_{\tilde{\imath}}|$ test items in total. On these items, compute Acc, Recall, and F1 with Fake as the positive class:
    \begin{equation*}
    \begin{aligned}
    \mathrm{Acc}_{\tilde{\imath}}
      &= \frac{\mathrm{TP}_{\tilde{\imath}} + \mathrm{TN}_{\tilde{\imath}}}{2|\mathcal{P}_{\tilde{\imath}}|},\\
    \mathrm{R}_{\tilde{\imath}}
      &= \frac{\mathrm{TP}_{\tilde{\imath}}}{\mathrm{TP}_{\tilde{\imath}} + \mathrm{FN}_{\tilde{\imath}}},\\
    \mathrm{F1}_{\tilde{\imath}}
      &= \frac{2\,\mathrm{TP}_{\tilde{\imath}}}{2\,\mathrm{TP}_{\tilde{\imath}} + \mathrm{FP}_{\tilde{\imath}} + \mathrm{FN}_{\tilde{\imath}}}.
    \end{aligned}
    \end{equation*}
    \item Take the mean across the $19$ generators (every generator gets the same weight):
    \begin{equation*}
    \begin{aligned}
    \mathrm{Acc}
      &= \tfrac{1}{|\mathcal{G}_{\text{fake}}|} \sum_{\tilde{\imath}} \mathrm{Acc}_{\tilde{\imath}},\\
    \mathrm{R}
      &= \tfrac{1}{|\mathcal{G}_{\text{fake}}|} \sum_{\tilde{\imath}} \mathrm{R}_{\tilde{\imath}},\\
    \mathrm{F1}
      &= \tfrac{1}{|\mathcal{G}_{\text{fake}}|} \sum_{\tilde{\imath}} \mathrm{F1}_{\tilde{\imath}}.
    \end{aligned}
    \end{equation*}
\end{enumerate}

\subsection{Additional Details of GenVideo}
\label{app:genvideo}

\noindent\textbf{Many to Many Zero Shot Eval.} GenVideo~\cite{genvideo2024} is the standard large scale benchmark for cross generator AI video detection. It has about $1{,}078{,}838$ fake videos and about $1{,}223{,}511$ real videos in total. We follow the \emph{many to many} test setup of GenVideo~\cite{genvideo2024}: the test set has about $18{,}588$ videos ($10{,}000$ real $+$ $8{,}588$ fake). The real videos come from MSR-VTT, Youku-mPLUG, and Kinetics-400, and form a shared pool that is reused for every test generator. The fakes are from $10$ generators (Sora, Morph Studio, Gen2, HotShot, Lavie, Show-1, MoonValley, Crafter, ModelScope, and WildScrape). The point of this setup is to check how well the model works on generators it has \emph{not} seen during training, not how well it works on the same generators. We report Recall and F1 on each generator and the average across generators, the same as in~\cite{genvideo2024,vidguard2025}.

\noindent\textbf{Benchmark-specific training data.} For the GenVideo result in Table~\ref{tab:genvideo} in the main paper, we train on the GenVideo training split, which has $10$ generators: Pika, ZeroScope, DynamiCrafter, VideoCrafter, SVD, OpenSora, Latte, SEINE, I2VGen-XL, and Stable Diffusion. These $10$ training generators are different from the $10$ test generators listed above, so the GenVideo many to many evaluation is fully zero shot.

\noindent\textbf{Evaluation metrics of GenVideo.} For this dataset, we follow GenVideo~\cite{genvideo2024} and VidGuard-R1~\cite{vidguard2025} and report Recall (R) and F1 with \textbf{Fake} as the positive class. The eval set per generator is $\mathcal{V}_{\tilde{\imath}} = \mathcal{V}_{\tilde{\imath}}^{\text{fake}} \cup \mathcal{V}^{\text{real}}$, where the shared real pool $\mathcal{V}^{\text{real}}$ is the same $\approx 10{,}000$ real videos described in Section~\ref{app:genvideo}, reused for every $\tilde{\imath}$. For each test generator $\tilde{\imath}$ ($|\mathcal{G}_{\text{fake}}| = 10$):
\begin{equation*}
\mathrm{R}_{\tilde{\imath}} = \frac{\mathrm{TP}_{\tilde{\imath}}}{|\mathcal{V}_{\tilde{\imath}}^{\text{fake}}|}, \qquad
\mathrm{F1}_{\tilde{\imath}} = \frac{2\,\mathrm{TP}_{\tilde{\imath}}}{2\,\mathrm{TP}_{\tilde{\imath}} + \mathrm{FP}_{\tilde{\imath}} + \mathrm{FN}_{\tilde{\imath}}}.
\end{equation*}
The final scores are the mean across the $10$ generators:
\begin{equation*}
\mathrm{R} = \tfrac{1}{|\mathcal{G}_{\text{fake}}|} \sum_{\tilde{\imath}} \mathrm{R}_{\tilde{\imath}}, \qquad
\mathrm{F1} = \tfrac{1}{|\mathcal{G}_{\text{fake}}|} \sum_{\tilde{\imath}} \mathrm{F1}_{\tilde{\imath}}.
\end{equation*}

\subsection{Additional Details of Eq.~\ref{eq:ridge} in the Main Paper}
\label{app:eq_ridge_details}
\label{app:influence}

This subsection derives how the classical influence function in Eq.~\ref{eq:inf} in the main paper is applied to our masked SFT objective and then reformulated as a ridge regression problem in Eq.~\ref{eq:ridge} in the main paper under a damped empirical Fisher approximation. The validation split and the conversion from raw influence scores to the sample weights in Eq.~\ref{eq:sft_loss} in the main paper are given in Section~\ref{app:eq_sft_loss_details}.

Our estimator builds on the ridge-regression reformulation and normalized stochastic solver of RRInf~\cite{rrinf2025}. We adapt this pipeline to the mask-consistent SFT objective for teacher-generated tool-use trajectories and convert the resulting signed influence estimates into weights for iterative trajectory training.

\noindent\textbf{Approximation roadmap.} The derivation below has both exact algebraic steps and implementation approximations. First, we apply the classical influence definition to the same masked token level objective used by SFT. Second, we approximate the Hessian with a damped empirical Fisher built from masked per sample gradients. Third, given that approximation, the push through identity exactly converts the damped inverse form into a ridge problem over $J_b$ sample coefficients. Fourth, the implementation solves this ridge problem with a normalised stochastic gradient method, where each mini batch is a small subset of trainable parameter coordinates and a LoRA layer is used as a structured coordinate block for memory efficiency. Finally, the influence weight normalisation rule converts the resulting signed estimated influence values into the weights used by the second SFT pass.

\noindent\textbf{From influence functions to ridge regression.}\label{app:if_derivation_full} This subsection has three parts. The first part is the math: how the classical influence function, when evaluated on the same masked token level objective as SFT, reduces to the ridge regression problem in the paragraphs from \ref{app:if_setup} to \ref{app:if_ridge}. The second part is the wall clock cost of running the resulting pipeline at the LoRA scale of Section~\ref{sec:stage1} (\ref{app:if_overhead}). The third part explains why the new form is cheap to run (\ref{app:if_complexity} and its three sub points). 

\noindent\textbf{LoRA restricted parameter space and mask consistent per sample gradients.}\label{app:if_setup} In Stage~1 we adapt the base multimodal model with LoRA. Each adapted weight matrix $W \in \mathbb{R}^{p \times q}$ is written as $W = W_0 + \frac{\alpha}{r} B A$, where $W_0$ is frozen, $\alpha$ is the LoRA scaling factor, and only the small factors $B \in \mathbb{R}^{p \times r}$, $A \in \mathbb{R}^{r \times q}$ ($r \ll \min(p, q)$) are trained. We put the LoRA factors of every adapted module together into one vector $\vartheta \in \mathbb{R}^{P_\vartheta}$. The size $P_\vartheta$ is much smaller than the full backbone. All gradients, Hessians and influence values below are computed only with respect to $\vartheta$. As is standard in influence analyses~\cite{datainf2023}, each influence computation is evaluated at the current adapter checkpoint, denoted by $\vartheta^{\star}$ for notational simplicity.

Besides, in our setting, each trajectory is only partially supervised. The student should learn the assistant side reasoning, answers, and tool call names, but system message and tool output are not generated by the student and are assigned label $-100$ in the SFT data (Section~\ref{app:label_masking}). If influence scores were computed on the unmasked transcript, they would measure a different objective from the one actually optimised by SFT, and could reward samples for fitting deterministic tool output text rather than for providing useful supervised reasoning and tool use decisions. Therefore, we define all influence quantities below using the same token level mask as Eq.~\ref{eq:naive_sft} in the main paper.

In the main text, we write the Hessian generically as $\mathcal{H}_\vartheta$; in the derivation below, we evaluate it at the current adapter checkpoint $\vartheta^\star$ and therefore write $\mathcal{H}_{\vartheta^\star}$.
For a training sample $z_j = (v_j, \tau_j^*)$, the per sample loss used for influence estimation is
\begin{equation}
\mathcal{L}_j(\vartheta)
= -\sum_{\eta} m_{\eta}^{(j)} \log p_{\vartheta}\!\left(\sigma_{\eta}^{(j)} \mid v_j, \widehat{\mathcal{I}}, \sigma_{<\eta}^{(j)}\right),
\label{eq:masked_loss_app}
\end{equation}
where $\{\sigma_{\eta}^{(j)}\}_{\eta}$ are the tokens of $\tau_j^*$ and $m_{\eta}^{(j)} \in \{0,1\}$ is the active token mask: $m_{\eta}^{(j)}=1$ for supervised assistant tokens and $m_{\eta}^{(j)}=0$ for tokens whose label is $-100$. We keep the notation $\mathcal{L}_j$ for this masked loss throughout the derivation. The per sample gradient is $g_j = \nabla_{\vartheta} \mathcal{L}_j(\vartheta^{\star}) \in \mathbb{R}^{P_\vartheta}$, and for a ridge batch of $J_b$ retained trajectories we stack the gradients into $\Psi = [\,g_1, \dots, g_{J_b}\,] \in \mathbb{R}^{P_\vartheta \times J_b}$. Here $J_b$ denotes a generic ridge batch size; the derivation allows $J_b \le J$, while in all reported experiments we use the full training split as the ridge batch, so $J_b=J=|\mathcal{S}_{\text{ref}}^{\text{train}}|=3{,}632$. Let $\widetilde{\mathcal{V}}_{\mathrm{IF}} = \{z_{\mathrm{val}}^{(1)}, \dots, z_{\mathrm{val}}^{(\widetilde V)}\}$ be the held out validation set (built in Section~\ref{app:if_val_split}). The averaged validation gradient is $\bar{b} = \frac{1}{\widetilde V}\sum_{h=1}^{\widetilde V} \nabla_{\vartheta} \mathcal{L}\!\left(\vartheta^{\star};\, z_{\mathrm{val}}^{(h)}\right) \in \mathbb{R}^{P_\vartheta}$, where $\mathcal{L}$ denotes the same masked loss as Eq.~\ref{eq:masked_loss_app}. The validation target $b$ used in the ridge regression solver below is a rescaled version of $\bar{b}$, defined in Eq.~\ref{eq:b_def_app}.

\noindent\textbf{Hessian simplification via Bartlett's second identity.}\label{app:if_hessian} Our starting point is the classical influence function~\cite{datainf2023} applied to the masked training and validation losses above:
\begin{equation}
\chi_j \;=\; -\,\bar{b}^{\top}\, \mathcal{H}_{\vartheta^{\star}}^{-1}\, g_j,
\qquad \mathcal{H}_{\vartheta^{\star}} \;=\; \nabla_{\vartheta}^{2} \mathcal{L}_\text{train}(\vartheta^{\star}),
\label{eq:if_classical_app}
\end{equation}
where $\mathcal{L}_\text{train}(\vartheta) := \frac{1}{J_b}\sum_{j=1}^{J_b} \mathcal{L}_j(\vartheta)$ is the \emph{averaged masked empirical risk} for the current ridge batch (the per sample mean version of the unnormalised masked sum in Eq.~\ref{eq:naive_sft} in the main paper); this normalisation is the standard convention in the influence function literature~\cite{datainf2023,rrinf2025} and is what makes the empirical Fisher in Eq.~\ref{eq:fisher_approx_app} consistent with $\nabla_\vartheta^2 \mathcal{L}_\text{train}$. Eq.~\ref{eq:if_classical_app} is intractable when $\mathcal{H}_{\vartheta^{\star}} \in \mathbb{R}^{P_\vartheta \times P_\vartheta}$ is dense. Throughout this derivation, identity matrices are written with explicit dimensions, such as $\mathbf{I}_{P_\vartheta}$ or $\mathbf{I}_{J_b}$. A standard identity for log likelihood losses, often referred to as Bartlett's second identity~\cite{datainf2023}, allows us to bypass second order computation entirely. Concretely, writing the score as $\widetilde{s}_{\vartheta}(c_{\mathrm{ctx}}, o) = \nabla_{\vartheta} \log p_{\vartheta}(o \mid c_{\mathrm{ctx}})$ for a generic conditioning context $c_{\mathrm{ctx}}$ and output token $o$, then under mild regularity:
\begin{equation}
\mathbb{E}\!\left[\widetilde{s}_{\vartheta}(c_{\mathrm{ctx}}, o)\, \widetilde{s}_{\vartheta}(c_{\mathrm{ctx}}, o)^{\top}\right] \;=\; -\,\mathbb{E}\!\left[\nabla_{\vartheta}^{2} \log p_{\vartheta}(o \mid c_{\mathrm{ctx}})\right],
\label{eq:bartlett2_app}
\end{equation}
where the expectation is taken over $(c_{\mathrm{ctx}}, o) \sim p_{\vartheta}$. Applied to our masked cross entropy training loss, Eq.~\ref{eq:bartlett2_app} suggests that, in expectation under the model distribution and at the current adapter checkpoint $\vartheta^{\star}$, the Hessian can be approximated by the outer product of the masked per sample gradients. The masks are fixed with respect to $\vartheta$, so the same first order gradients $g_j$ in Eq.~\ref{eq:masked_loss_app} define both the empirical Fisher approximation and the ridge problem below. We do not invoke Bartlett's identity as an exact equality (our SFT samples are teacher trajectories, not draws from $p_{\vartheta}$); it merely motivates the standard \emph{empirical Fisher} approximation~\cite{datainf2023,rrinf2025} that replaces the Hessian by the corresponding sample average:
\begin{equation}
\mathcal{H}_{\vartheta^{\star}} \;\approx\; \widetilde{\mathcal{H}} \;:=\; \frac{1}{J_b}\sum_{j=1}^{J_b} g_j\, g_j^{\top} \;=\; \frac{1}{J_b}\,\Psi\,\Psi^{\top} \;\in\; \mathbb{R}^{P_\vartheta \times P_\vartheta}.
\label{eq:fisher_approx_app}
\end{equation}
Eq.~\ref{eq:fisher_approx_app} is good for our setting in two ways. First, we do not need a second order backward pass; we only need first order per sample gradients, which we obtain with one extra forward/backward pass per training sample over the trainable LoRA factors only (not a free byproduct of mini batch SFT, which yields only batch aggregated gradients). Second, $\widetilde{\mathcal{H}}$ is positive semidefinite by construction. So once we add a small ridge term for numerical stability, the inverse is well defined.

\noindent\textbf{Reduction to a ridge regression problem.}\label{app:if_ridge} Substituting Eq.~\ref{eq:fisher_approx_app} into a damped version of Eq.~\ref{eq:if_classical_app} with Hessian damping coefficient $\rho_{\mathcal{H}} > 0$, we define the damped empirical-Fisher influence estimate as:
\begin{equation}
\begin{aligned}
\widehat{\chi}_j
  &:= -\,\bar{b}^{\top}\!\left(\widetilde{\mathcal{H}}
      + \rho_{\mathcal{H}} \mathbf{I}_{P_\vartheta}\right)^{-1} g_j\\
  &= -\,J_b \cdot \bar{b}^{\top}\!\left(\Psi\, \Psi^{\top}
      + J_b\rho_{\mathcal{H}} \mathbf{I}_{P_\vartheta}\right)^{-1} g_j .
\end{aligned}
\label{eq:if_fisher_app}
\end{equation}
For the ridge formulation, we absorb the leading $-J_b$ factor by defining the validation target as
\begin{equation}
b \;:=\; -\,J_b\,\bar{b} \;=\; -\,\frac{J_b}{\widetilde V}\sum_{h=1}^{\widetilde V} \nabla_{\vartheta} \mathcal{L}\!\left(\vartheta^{\star};\, z_{\mathrm{val}}^{(h)}\right) \;\in\; \mathbb{R}^{P_\vartheta} ,
\label{eq:b_def_app}
\end{equation}
so that Eq.~\ref{eq:if_fisher_app} simplifies to $\widehat{\chi}_j = b^{\top}(\Psi\,\Psi^{\top} + J_b\rho_{\mathcal{H}} \mathbf{I}_{P_\vartheta})^{-1} g_j$, with the leading scale fully absorbed into $b$.
Computing this expression directly still needs the inverse of a $P_\vartheta \times P_\vartheta$ matrix, which is too large at our LoRA scale. We use the push through reformulation to move the inverse to the much smaller $J_b \times J_b$ space:
\begin{equation}
\left(\Psi\, \Psi^{\top} + \mu \mathbf{I}_{P_\vartheta}\right)^{-1} \Psi \;=\; \Psi \left(\Psi^{\top} \Psi + \mu \mathbf{I}_{J_b}\right)^{-1}, \qquad \forall\, \mu > 0,
\label{eq:pushthrough_app}
\end{equation}
where $\mu$ is a generic positive constant. This identity is a direct corollary of the Sherman--Morrison--Woodbury identity. Plugging Eq.~\ref{eq:pushthrough_app} into the simplified form of Eq.~\ref{eq:if_fisher_app} (with $\mu = J_b\rho_{\mathcal{H}}$, and using the absorbed $b$ from Eq.~\ref{eq:b_def_app}) gives the per sample damped empirical-Fisher influence estimate vector
\begin{equation}
\bigl[\widehat{\chi}_1,\, \dots,\, \widehat{\chi}_{J_b}\bigr]^{\top} \;=\; \left(\Psi^{\top} \Psi + J_b\rho_{\mathcal{H}} \mathbf{I}_{J_b}\right)^{-1} \Psi^{\top} b.
\label{eq:if_kernel_app}
\end{equation}
The right hand side has no residual scale: the $-J_b$ factor that would otherwise sit outside has been absorbed into $b$ at definition time (Eq.~\ref{eq:b_def_app}).

\noindent\textbf{Equivalence to ridge regression.} The right hand side of Eq.~\ref{eq:if_kernel_app} is exactly the closed form solution of the corresponding damped empirical-Fisher influence estimate as a ridge regression problem in $\mathbb{R}^{J_b}$ with ridge penalty $\rho > 0$ (this is the $\rho$ that appears in Eq.~\ref{eq:ridge} in the main paper):
\begin{equation}
\hat{\xi} \;=\; \operatorname*{arg\,min}_{\xi \in \mathbb{R}^{J_b}} \;\; \frac{1}{P_\vartheta}\,\bigl\|\Psi\,\xi - b\bigr\|_{2}^{2} \;+\; \rho\,\bigl\|\xi\bigr\|_{2}^{2},
\label{eq:if_ridge_app}
\end{equation}
We can check this by setting the gradient of the objective in Eq.~\ref{eq:if_ridge_app} to zero:
\begin{equation}
\begin{aligned}
\frac{2}{P_\vartheta}\,\Psi^{\top}\!\bigl(\Psi\,\hat{\xi} - b\bigr)
  + 2\rho\,\hat{\xi} &= 0\\
\Longleftrightarrow\qquad
\hat{\xi}
  &= \left(\Psi^{\top}\Psi + P_\vartheta\rho \mathbf{I}_{J_b}\right)^{-1}
     \Psi^{\top} b.
\end{aligned}
\end{equation}
The Hessian damping $\rho_{\mathcal{H}}$ (Eq.~\ref{eq:if_fisher_app}) and the ridge penalty $\rho$ (Eq.~\ref{eq:if_ridge_app}) play different roles, and they relate by a $J_b/P_\vartheta$ scaling: $\widehat{\chi}_j = \hat{\xi}_j$ holds exactly when $\rho = (J_b/P_\vartheta)\,\rho_{\mathcal{H}}$. In practice, however, we do \emph{not} pick a flat numerical value for either $\rho_{\mathcal{H}}$ or $\rho$. Instead, we use an adaptive layer wise damping parameterisation. For LoRA layer $a$, let $P_a$ be the number of trainable adapter parameters in that layer, let $\Psi^{(a)} = [g_1^{(a)}, \dots, g_{J_b}^{(a)}] \in \mathbb{R}^{P_a \times J_b}$ be the corresponding gradient slab, and define $\overline{\|g^{(a)}\|^2} := J_b^{-1}\sum_{j=1}^{J_b}\|g_j^{(a)}\|_2^2$ as the average squared per sample gradient norm over the ridge batch. The per layer Hessian damping is set as
\begin{equation}
\rho_{\mathcal{H}}^{(a)} \;=\; \frac{1}{\kappa}\,\cdot\,\frac{1}{J_b\,P_a}\,\bigl\|\Psi^{(a)}\bigr\|_F^2
\;=\;\frac{1}{\kappa}\,\cdot\,\overline{\|g^{(a)}\|^2}\,/\,P_a,
\label{eq:rho_adaptive_app}
\end{equation}
where $\kappa > 0$ is the adaptive damping coefficient, i.e.\ the user level scaling knob in our implementation. Using the same relation, the associated layer wise effective ridge penalty is $\rho^{(a)} = (J_b/P_\vartheta)\,\rho_{\mathcal{H}}^{(a)}$, where $P_\vartheta$ is the full trainable gradient dimension; the layer size $P_a$ only enters the adaptive damping estimate in Eq.~\ref{eq:rho_adaptive_app}. Thus the effective regularisation is data dependent and varies across layers, rather than being a flat $\rho_{\mathcal{H}} = 10$ or a flat $\rho = 10$. Throughout the paper, we reserve $\rho_{\mathcal{H}}$ for Hessian damping and $\rho$ for the ridge penalty in Eq.~\ref{eq:if_ridge_app}; our implementation does not set either one to a single flat numerical value directly, and instead uses the adaptive damping coefficient $\kappa$ to determine the layer wise effective regularisation through Eq.~\ref{eq:rho_adaptive_app}. In the full LoRA-gradient case, setting $\widehat{\chi}_j = \hat{\xi}_j$ recovers
Eq.~\ref{eq:ridge} in the main paper; the layer-wise stochastic solver below is used as a practical approximation to this ridge objective.

\noindent\textbf{Why the ridge form matters.} Eq.~\ref{eq:if_ridge_app} replaces a hard $P_\vartheta \times P_\vartheta$ inverse with a strongly convex quadratic program in $J_b$ unknowns. The dense ridge objective admits simple gradient descent updates, where each full-gradient iteration costs one matrix vector product. As stated above, in all reported experiments the ridge batch covers the full training split, so $J_b=J=3{,}632$. For the LoRA setup in Section~\ref{sec:stage1}, $P_\vartheta \approx 8.07 \times 10^{7}$ and $J_b = 3{,}632$, so the unknown dimension drops from $P_\vartheta$ to $J_b$, which is a roughly $2.2 \times 10^4$-fold reduction ($P_\vartheta/J_b$); equivalently, the dense matrix shrinks from $P_\vartheta \times P_\vartheta$ to $J_b \times J_b$, about $5 \times 10^8$ \emph{times} fewer entries (the absolute drop is $\sim 6.5 \times 10^{15}$ entries).

\noindent\textbf{Why the ridge regression pipeline is cheap.}\label{app:if_complexity} Three properties together make this pipeline cheap. \emph{(i) One backward pass per sample.}
\label{app:if_phi_cost}
The first stage builds the masked gradient matrix $\Psi = [g_1, \dots, g_{J_b}] \in \mathbb{R}^{P_\vartheta \times J_b}$. Each mask consistent per sample gradient $g_j$ needs only one forward pass and one backward pass through the network, with system, and tool output tokens excluded by Eq.~\ref{eq:masked_loss_app}. The frozen backbone is still traversed in both directions, but only the LoRA factors carry trainable gradients (so storage and the optimiser update touch only those), and we do not need any second order computation. So building $\Psi$ for a ridge batch of $J_b$ samples costs about the same as one extra epoch of LoRA forward and backward passes when the batch covers the training part. \emph{(ii) Push through identity removes the $P_\vartheta \times P_\vartheta$ inverse.}
\label{app:if_solver_cost}
As shown in Eq.~\ref{eq:pushthrough_app}, the push through identity changes the $P_\vartheta \times P_\vartheta$ inverse of the empirical Fisher into a $J_b \times J_b$ inverse of the Gram matrix $\Psi^{\top}\Psi + P_\vartheta\rho \mathbf{I}_{J_b}$. The closed form complexity drops from $\mathcal{O}(P_\vartheta^{3})$ FLOPs to $\mathcal{O}(J_b^{3} + P_\vartheta J_b^{2})$ FLOPs. To put the numbers in our setting: the $\mathcal{O}(P_\vartheta^{3})$ form is plainly out of reach at $P_\vartheta \approx 8.07 \times 10^{7}$; the Gram matrix itself is small ($J_b^2 \approx 1.3 \times 10^{7}$ entries), but \emph{forming} it costs $\mathcal{O}(P_\vartheta J_b^2) \sim 10^{15}$ FLOPs, and just holding $\Psi$ in memory already takes $\mathcal{O}(P_\vartheta J_b) \sim 3 \times 10^{11}$ entries. What actually makes the pipeline practical is a stochastic gradient method that only uses a small subset of model parameters per iteration. Let $\psi_{\varpi}^{\top} \in \mathbb{R}^{1 \times J_b}$ be the $\varpi$-th row of $\Psi$ and let $b_{\varpi}$ be the corresponding coordinate of $b$. The full-gradient ridge update can be written as
\begin{equation}
\begin{aligned}
\xi^{(t+1)}
  &= \xi^{(t)} - \nu_{\mathrm{sg}}\!\Biggl[
     \frac{2}{P_\vartheta}\sum_{\varpi=1}^{P_\vartheta}\psi_{\varpi}\\[-0.2em]
  &\hspace{7em}\bigl(\psi_{\varpi}^{\top}\xi^{(t)}-b_{\varpi}\bigr)
     +2\rho\,\xi^{(t)}\Biggr].
\end{aligned}
\end{equation}
Instead of summing over all $P_\vartheta$ coordinates, stochastic gradient descent uses a coordinate mini batch $C_t \subset \{1,\dots,P_\vartheta\}$:
\begin{equation}
\begin{aligned}
\xi^{(t+1)}
  &= \xi^{(t)} - \nu_{\mathrm{sg}}\!\Biggl[
     \frac{2}{|C_t|}\sum_{\varpi\in C_t}\psi_{\varpi}\\[-0.2em]
  &\hspace{7em}\bigl(\psi_{\varpi}^{\top}\xi^{(t)}-b_{\varpi}\bigr)
     +2\rho\,\xi^{(t)}\Biggr].
\end{aligned}
\end{equation}
Following RRInf~\cite{rrinf2025}, we further normalise each coordinate contribution by its squared gradient norm:
\begin{equation}
\begin{aligned}
\xi^{(t+1)}
  &= \xi^{(t)} - \nu_{\mathrm{sg}}\!\Biggl[
     \frac{2}{|C_t|}\sum_{\varpi\in C_t}
     \frac{\psi_{\varpi}}{\|\psi_{\varpi}\|_2^2+\epsilon}\\[-0.2em]
  &\hspace{7em}\bigl(\psi_{\varpi}^{\top}\xi^{(t)}-b_{\varpi}\bigr)
     +2\rho\,\xi^{(t)}\Biggr],
\end{aligned}
\end{equation}
where $\epsilon>0$ is a small numerical constant. In our implementation, $\nu_{\mathrm{sg}}=0.01$, $T_{\mathrm{ridge}} = 1{,}000$, and $\rho$ takes the adaptive value induced by Eq.~\ref{eq:rho_adaptive_app} with adaptive damping coefficient $\kappa = 10$. \emph{(iii) Layer wise sampling.}
\label{app:if_total_cost}
For memory efficiency, we instantiate $C_t$ as a structured coordinate mini batch corresponding to one LoRA layer. Concretely, let $\Lambda_a$ be the set of trainable coordinates in LoRA layer $a$, with $|\Lambda_a|=P_a$. We draw a layer with probability $q_a=P_a/P_\vartheta$ and set $C_t=\Lambda_{a_t}$, which is equivalent to drawing a trainable coordinate uniformly and then loading the full layer containing it. This keeps the stochastic update aligned with coordinate mini batching while exploiting the fact that adapter gradients are naturally stored by layer. The peak memory drops from $\mathcal{O}(P_\vartheta J_b)$ (full gradient) to $\mathcal{O}(\max_{a} P_a \cdot J_b)$ (one layer slab on device at a time), and the per step compute drops from $\mathcal{O}(P_\vartheta J_b)$ to $\mathcal{O}(P_{a_t} J_b)$ for the sampled layer $a_t$, with expected per step compute $\mathcal{O}(\mathbb{E}_{a\sim q}[P_a] \cdot J_b)$. Total compute over $T_{\mathrm{ridge}}$ iterations is therefore $\mathcal{O}(T_{\mathrm{ridge}} P_\vartheta J_b)$ for the full-gradient solver and $\mathcal{O}(T_{\mathrm{ridge}}\,\mathbb{E}_{a\sim q}[P_a]\,J_b)$ for the layer wise variant. Together, these three properties make the whole IF pipeline only a constant factor slower than naive LoRA SFT. There is no second order computation, no $P_\vartheta \times P_\vartheta$ inverse, and the working memory is bounded by the gradients of one LoRA layer. This explains why the ridge regression pipeline is cheap.

\subsection{Additional Details of Eq.~\ref{eq:sft_loss} in the Main Paper}
\label{app:eq_sft_loss_details}

\noindent\textbf{Validation set construction.}\label{app:if_val_split} The validation set $\widetilde{\mathcal{V}}_{\mathrm{IF}}$ that we use to compute the masked validation target $b = -\frac{J_b}{\widetilde V}\sum_{h=1}^{\widetilde V} \nabla_\vartheta \mathcal{L}(\vartheta^\star; z_{\mathrm{val}}^{(h)})$ (Eq.~\ref{eq:b_def_app}) is a held out subset of the retained reference set $\mathcal{S}_{\text{ref}}$ introduced in Section~\ref{sec:stage1}, where $\mathcal{L}$ is the same masked objective defined in Eq.~\ref{eq:masked_loss_app}. Before any influence computation, we construct $\widetilde{\mathcal{V}}_{\mathrm{IF}}$ by sampling $10\%$ of the fake videos from each video generator, yielding $201$ fake videos in total, and then adding their corresponding $201$ paired real videos. Thus, the validation set contains $\widetilde V = |\widetilde{\mathcal{V}}_{\mathrm{IF}}| = 402$ videos. The remaining $J = |\mathcal{S}_{\text{ref}}^{\text{train}}| = 3{,}632$ videos form the training split $\mathcal{S}_{\text{ref}}^{\text{train}} = \mathcal{S}_{\text{ref}} \setminus \widetilde{\mathcal{V}}_{\mathrm{IF}}$. We compute the validation target $b$ only on $\widetilde{\mathcal{V}}_{\mathrm{IF}}$ and compute the per sample training gradients only on $\mathcal{S}_{\text{ref}}^{\text{train}}$, so the same held out target is used throughout influence reweighting.

\noindent\textbf{Influence weight normalisation.}\label{app:if_normalize} This subsection shows how to turn the raw estimated scores $\{\widehat{\chi}_j\}_{j=1}^{J}$ from Eq.~\ref{eq:if_ridge_app} into the per sample weights $\{w_j\}_{j=1}^{J}$ in the reweighted SFT loss of Eq.~\ref{eq:sft_loss} in the main paper.

\noindent\textbf{Sign convention and utility score.} As discussed below Eq.~\ref{eq:inf} in the main paper, the classical influence function convention assigns \emph{negative} scores to high quality samples (up weighting them reduces validation loss). With the absorbed validation target $b$ from Eq.~\ref{eq:b_def_app}, the ridge solution $\hat{\xi}$ inherits this convention exactly for the damped empirical-Fisher estimate: $\hat{\xi}_j = \widehat{\chi}_j < 0$ for high quality samples. To work with the ``higher is better'' convention used by the normalisation rule below, we introduce an explicit \emph{utility score}
\begin{equation}
u_j \;:=\; -\,\widehat{\chi}_j \;=\; -\,\hat{\xi}_j ,
\label{eq:utility_app}
\end{equation}
so that $u_j > 0$ means sample $j$ is good for training and $u_j \le 0$ means it is not. The estimated score $\widehat{\chi}_j$ follows the same sign convention as the original classical influence score $\chi_j$; only $u_j$ is used in the weighting rule.

Given the utility scores $\{u_j\}_{j=1}^{J}$, we use a simple influence weight normalisation rule with no hyperparameter:
\begin{equation}
w_j \;=\; J \cdot \frac{\max\!\bigl(u_j,\, 0\bigr)}{\sum_{k=1}^{J} \max\!\bigl(u_k,\, 0\bigr)}.
\label{eq:hard_drop_neg}
\end{equation}

\noindent\textbf{Properties.} Eq.~\ref{eq:hard_drop_neg} has three properties that we like in practice:
\begin{itemize}[nosep,leftmargin=*]
    \item \emph{Loss scale stays the same.} By design, $\sum_{j=1}^{J} w_j = J$. So the average per sample loss in the reweighted SFT loss of Eq.~\ref{eq:sft_loss} in the main paper matches that of the unweighted loss in Eq.~\ref{eq:naive_sft} in the main paper. This means the effective learning rate does not change, and we can reuse the optimiser settings of the first (unweighted) SFT pass without retuning.
    \item \emph{Hard filter of bad samples.} Any sample with $u_j \leq 0$ (equivalently $\widehat{\chi}_j \geq 0$) gets $w_j = 0$, so it is fully removed from the second SFT pass, not just down weighted. This matches our goal: low quality reference trajectories should not affect the loss at all.
    \item \emph{No hyperparameter.} Eq.~\ref{eq:hard_drop_neg} has no temperature, threshold, or scale knob. The only knob is the adaptive damping coefficient $\kappa$ (Eq.~\ref{eq:rho_adaptive_app}), which we already set in the influence solver. This is on purpose: in the ablations of Section~\ref{sec:exp}, any gap between the unweighted and the weighted SFT can come only from the influence based filter in Eq.~\ref{eq:hard_drop_neg}, not from some extra weighting knob.
\end{itemize}

\noindent\textbf{Edge case.} If $\sum_{k} \max(u_k, 0) \le 0$ (that is, no sample is good under the current ridge solution), we fall back to uniform weights $w_j = 1$ for all $j$. This is the same as the unweighted SFT loss in Eq.~\ref{eq:naive_sft} in the main paper. We have not seen this fallback fire in any of our runs.

\section{Toolset Details}
\label{app:toolset_details}
\label{app:toolset}

\noindent\textbf{Initial heuristic toolset.}\label{app:tool_registry} Table~\ref{tab:full_tool_registry} lists every tool in the initial toolset.Every round of code tool returns a short structured XML string (Section~\ref{app:tool_io}).

\begin{table*}[t]
\centering
\caption{Initial code tool registry.}
\label{tab:full_tool_registry}
\small
\setlength{\tabcolsep}{4pt}
\begin{tabular}{p{6.8cm}p{1.2cm}p{5.4cm}}
\toprule
Name & Class & Purpose (one line) \\
\midrule
\texttt{halo\_peak\_analyzer}                       & code  & Detect halo / ringing peaks around high gradient boundaries via halo peak statistics. \\
\texttt{halo\_spectrum\_analyzer}                   & code  & Detect halo with broader spectrum aware criteria (purple channel area). \\
\texttt{color\_cast\_analyzer}                      & code  & Detect frame to frame R-G channel jitter paired with oversaturated rendering. \\
\texttt{first\_frame\_jump\_analyzer}               & code  & Detect anomalous first frame discontinuity via robust MAD-z on bright and sharp channels. \\
\texttt{subtle\_jump\_analyzer}                     & code  & Detect lag-1 / lag-2 first frame jumps in the moderate MAD-z band from 3 to 20. \\
\texttt{background\_motion\_analyzer}               & code  & Detect background point drift under a confidently static camera. \\
\texttt{foreground\_burst\_analyzer}                & code  & Detect foreground point teleportation under a confidently static camera. \\
\texttt{forensic\_frame\_delta\_probe}              & code  & Judge frame to frame consistency based on inter frame change dimensions (color fingerprint, hand anomaly, jump stable patches, OCR drop, $\dots$). \\
\texttt{neutral\_reconsider\_distinct\_fingers}     & code  & Perform a targeted second check for hand and finger artifacts. \\
\texttt{neutral\_reconsider\_distorted\_chars}      & code  & Perform a targeted second check for distorted characters. \\
\texttt{neutral\_reconsider\_merge\_into}           & code  & Perform a targeted second check for multiple objects merging into a single object. \\
\texttt{neutral\_reconsider\_not\_stable}           & code  & Perform a targeted second check for unstable object forms. \\
\midrule
\texttt{toolbox\_guide}                     & guide & toolbox manual: introduces every code tool and dispatches the appropriate combination. \\

\bottomrule
\end{tabular}
\end{table*}

\noindent\textbf{Tool Card in Prompt.}\label{app:tool_card} Here is the tool card we use during reference construction, model training and inference. We give this to teacher model or student model when we want to let them know what tools they can use and how to use them. During tool evolution, this block is updated whenever the toolset evolves. The card looks like this:
\begin{lstlisting}[basicstyle=\ttfamily\footnotesize]
- first_frame_jump_analyzer [code]
  purpose: Detect anomalous first-frame discontinuity via a robust
           MAD-z computed on the bright and sharp channels.
  reads:   state.bright (per-frame brightness, 16 floats)
           state.sharp  (per-frame Laplacian variance, 16 floats)
  output:  XML string with an <ANALYSIS> body; additional
           output metadata is omitted here (...).
  fires when: bright-channel z > 4 OR sharp-channel z > 4.
  cannot prove: A first-frame baseline shift that affects all 16
                frames uniformly cancels out and is invisible to z.
\end{lstlisting}

\noindent\textbf{Tool output schema.}\label{app:tool_io} The outputs of all called tools in one round will be packed into one tool output that the model reads as the format below.
\begin{lstlisting}[basicstyle=\ttfamily\footnotesize]
<ANALYSIS>
... a short body that names the specific signals the tools
checked, the numeric values computed, and which thresholds
were exceeded (or not) ...
</ANALYSIS>
<MANUAL>
... a short natural-language explanation that the tool registry
uses to decide whether and how to override its baseline verdict
when this tool fires; not surfaced to the student in raw form ...
</MANUAL>
\end{lstlisting}

\noindent\textbf{Representative tool source.}\label{app:tool_code} We let the large model observe training set samples, and prompt it with some common tools in the field of AI-generated video detection to enable it to build the initial heuristic toolset. We show two representative tool implementations below.

\noindent\textbf{(i) First frame jump analyser.} Detects an anomalous discontinuity at the first sampled frame. Both the bright and sharp channels are summarised by a robust MAD-z that compares the magnitude of $|F_1 - F_0|$ against the median absolute deviation of $|F_{i+1} - F_i|$ over the rest of the clip; if either channel exceeds the threshold, the first frame is statistically out of distribution relative to its own clip.

\begin{lstlisting}[language=Python,basicstyle=\ttfamily\footnotesize]
from typing import Any, Dict


Z_THR = 4.0


def _mad_z(values):
    """Robust MAD-z of |diff(values)[0]| against the tail baseline."""
    values = [float(x) for x in values if x is not None]
    if len(values) < 3:
        return 0.0
    diffs = [abs(values[i + 1] - values[i]) for i in range(len(values) - 1)]
    tail = diffs[1:]
    sorted_tail = sorted(tail)
    n = len(sorted_tail)
    base = sorted_tail[n // 2] if n % 2 else 0.5 * (sorted_tail[n // 2 - 1] + sorted_tail[n // 2])
    abs_dev = sorted(abs(x - base) for x in tail)
    m = len(abs_dev)
    mad = abs_dev[m // 2] if m % 2 else 0.5 * (abs_dev[m // 2 - 1] + abs_dev[m // 2])
    return (diffs[0] - base) / (1.4826 * mad + 1e-6)


def run(video_id: str, state: Dict[str, Any]) -> str:
    bright = state.get("bright") or []
    sharp = state.get("sharp") or []

    if len(bright) < 3 or len(sharp) < 3:
        return (
            "<ANALYSIS>\n"
            "First-frame analysis: insufficient frames.\n"
            "</ANALYSIS>\n"
            "..."
        )

    b_z = _mad_z(bright)
    s_z = _mad_z(sharp)

    if b_z > Z_THR or s_z > Z_THR:
        return (
            "<ANALYSIS>\n"
            f"First-frame robust MAD-z (vs frames 2-16 baseline):\n"
            f"  bright channel z = {b_z:.2f}, sharp channel z = {s_z:.2f}\n"
            f"  threshold {Z_THR:.1f} exceeded; first frame is\n"
            f"  out-of-distribution relative to the rest of the clip.\n"
            "</ANALYSIS>\n"
            "..."
        )

    return (
        "<ANALYSIS>\n"
        f"First-frame robust MAD-z (vs frames 2-16 baseline):\n"
        f"  bright channel z = {b_z:.2f}, sharp channel z = {s_z:.2f}\n"
        f"  threshold {Z_THR:.1f} not exceeded.\n"
        "</ANALYSIS>\n"
        "..."
    )
\end{lstlisting}

\noindent\textbf{(ii) Background motion analyser.} A physics grounded check: under a confidently static camera, tracked background points should not drift. The tool fires only when the camera motion classifier is confident the camera is static and there are enough background points for a reliable percentile estimate; it then asks whether the $90$th percentile of background point displacement and per frame step both exceed their thresholds.

\begin{lstlisting}[language=Python,basicstyle=\ttfamily\footnotesize]
from typing import Any, Dict


CAM_CONF_THR = 0.87
MIN_NBG = 10
DISP_P90_THR = 25.0
STEP_P90_THR = 8.0


def run(video_id: str, state: Dict[str, Any]) -> str:
    cam_type = state.get("cam_type")
    cam_conf = float(state.get("cam_conf") or 0.0)
    n_bg = int(state.get("n_bg") or 0)
    disp_p90 = state.get("disp_p90")
    step_p90 = state.get("step_p90")

    if cam_type != "static" or cam_conf < CAM_CONF_THR:
        return (
            "<ANALYSIS>\n"
            f"Background motion: camera is not confidently static "
            f"(type={cam_type}, conf={cam_conf:.2f}); rule skipped.\n"
            "</ANALYSIS>\n"
            "..."
        )
    if n_bg < MIN_NBG or disp_p90 is None or step_p90 is None:
        return (
            "<ANALYSIS>\n"
            f"Background motion: insufficient background points "
            f"(n_bg={n_bg}); rule skipped.\n"
            "</ANALYSIS>\n"
            "..."
        )

    disp_p90 = float(disp_p90)
    step_p90 = float(step_p90)
    if disp_p90 > DISP_P90_THR and step_p90 > STEP_P90_THR:
        return (
            "<ANALYSIS>\n"
            f"Background motion under static camera "
            f"(n_bg={n_bg}, cam_conf={cam_conf:.2f}):\n"
            f"  background-point displacement p90 = {disp_p90:.1f} px "
            f"(threshold {DISP_P90_THR:.0f})\n"
            f"  background-point per-frame step p90 = {step_p90:.1f} px "
            f"(threshold {STEP_P90_THR:.0f})\n"
            f"  Both thresholds exceeded; background drift is\n"
            f"  physically inconsistent with a static camera.\n"
            "</ANALYSIS>\n"
            "..."
        )

    return (
        "<ANALYSIS>\n"
        f"Background motion under static camera "
        f"(disp_p90={disp_p90:.1f}, step_p90={step_p90:.1f}); "
        f"thresholds not jointly exceeded.\n"
        "</ANALYSIS>\n"
        "..."
    )
\end{lstlisting}

\noindent\textbf{Initial heuristic toolset vs. updated toolset.}\label{app:C_lib_snapshot}\label{app:evo_snapshot} We compare the initial heuristic toolset and the updated toolset. The initial heuristic toolset has $13$ tools as listed in Table~\ref{tab:full_tool_registry}: $12$ code tools plus \texttt{toolbox\_guide}.

\begin{table*}[t]
\centering
\caption{Comparison between the initial heuristic toolset and updated toolset of our framework.}
\label{tab:lib_diff}
\footnotesize
\setlength{\tabcolsep}{6pt}
\begin{tabular}{p{0.45\textwidth}p{0.45\textwidth}}
\toprule
Initial heuristic toolset & Updated toolset \\
\midrule
\toolname{halo_peak_analyzer}                   & \toolname{halo_peak_analyzer} \\
\toolname{halo_spectrum_analyzer}               & \toolname{halo_spectrum_analyzer} \\
\toolname{color_cast_analyzer}                  & \toolname{color_cast_analyzer} \\
\toolname{first_frame_jump_analyzer}            & \toolname{first_frame_jump_analyzer} \\
\toolname{subtle_jump_analyzer}                 & \toolname{subtle_jump_analyzer} \\
\toolname{background_motion_analyzer}           & \toolname{background_motion_analyzer} \\
\toolname{foreground_burst_analyzer}            & \toolname{foreground_burst_analyzer} \\
\toolname{forensic_frame_delta_probe}           & \toolname{forensic_frame_delta_probe} \\
\toolname{neutral_reconsider_distinct_fingers}  & \toolname{neutral_reconsider_distinct_fingers} \\
\toolname{neutral_reconsider_distorted_chars}   & \toolname{neutral_reconsider_distorted_chars} \\
\toolname{neutral_reconsider_merge_into}        & \toolname{neutral_reconsider_merge_into} \\
\toolname{neutral_reconsider_not_stable}        & \toolname{neutral_reconsider_not_stable} \\
\toolname{toolbox_guide}                        & \toolname{toolbox_guide} \\
                                                & \toolname{protect_real_bg_density} \\
                                                & \toolname{camconf_low_fake_check} \\
                                                & \toolname{motion_step_fuser} \\
                                                & \toolname{complex_scene_consistency_fuser} \\
                                                & \toolname{blue_sky_foreground_density_check} \\
                                                & \toolname{face_landmark_consistency} \\
                                                & \toolname{sam_region_stability} \\
                                                & \toolname{ocr_sam_filter} \\
                                                & \toolname{physical_implausibility_analyzer} \\
\bottomrule
\end{tabular}
\end{table*}
\section{Prompts}
\label{app:supp_prompts}
\label{app:D_prompts}
\label{app:E_prompts}

In this section, we collect the prompts referenced earlier in the appendix. The teacher prompts are referenced from Section~\ref{app:teacher_prompt}, the student facing SFT/GRPO/evaluation prompt is listed in Section~\ref{app:D_student_prompt}, and the tool evolution operator prompts are referenced from Section~\ref{app:evo_prompts}.

\subsection{Teacher Prompts}
\label{app:D_teacher}

Below are two teacher prompts for both pipelines. After every round's teacher model's output, we run the invoked tools. After the tools return results, we append the tool output and previous messages to the end of this prompt, then send the combined content to the teacher model to generate the next round of output.

\noindent\textbf{GenVideo (generate-then-filter reference construction; GT is not included in the teacher prompt).}
\begin{lstlisting}[style=prompt]
Here are uniformly sampled frames in one video sample:
- frame_idx 0:  <attached image>
- frame_idx 1:  <attached image>
- ... 

Your task is to use the frames and the toolset provided later to reason about whether the video is AI-generated or real. Include <think>, an <answer>, 1-3 code-tool calls from the tool_card block, then toolbox_guide as the last tool call. Stop immediately after toolbox_guide. Reason with these tools and decide; call the tools from the tool cards below that you think can help you further inspect artifacts in the video. (tool calls refer to frames by sampled-frame index starting from 0). 

Below is a toolset, presented as tool cards, that helps you judge whether the given video sample is real or AI-generated.
<tool_cards>
... one compact card per forensic tool ...
</tool_cards>

schema of your output:
<think>...</think>
<answer>Real|Fake</answer>
You could also append with tool call as below:
<tool>tool_name_1</tool>
<tool>tool_name_2</tool>
...
<tool>toolbox_guide</tool>

General rules of your output:
- Only tools in the tool-card block may appear inside <tool> tags.
- Every <tool> block takes either the form <tool>{"name": str, "args":          
  dict}</tool> or <tool>{"name": str}</tool>, depending on the tool_card.
- The tool results is not produced by you. Do not fabricate any tool output.


\end{lstlisting}

\noindent\textbf{ViF-CoT-4K (rewrite of the official ViF-CoT-4K reference).}
\begin{lstlisting}[style=prompt]

Here are uniformly sampled frames in one video sample:
- frame_idx 0:  <attached image>
- frame_idx 1:  <attached image>
- ... 

Your task is to use the frames and the toolset provided later to reason about whether the video is AI-generated or real. Include <think>, an <answer>, 1-3 code-tool calls from the tool_card block, then toolbox_guide as the last tool call. Stop immediately after toolbox_guide. Reason with these tools and decide; call the tools from the tool cards below that you think can help you further inspect artifacts in the video. (tool calls refer to frames by sampled-frame index starting from 0). 

Below is a toolset, presented as tool cards, that helps you judge whether the given video sample is real or AI-generated.
<tool_cards>
... one compact card per forensic tool ...
</tool_cards>

The official ViF-CoT-4K reference reasoning for this sample is as
follows:
<vifcot4k_reference>

schema of your output:
<think>...</think>
<answer>Real|Fake</answer>
You could also append with tool call as below:
<tool>tool_name_1</tool>
<tool>tool_name_2</tool>
...
<tool>toolbox_guide</tool>

General rules of your output:
- Only tools in the tool-card block may appear inside <tool> tags.
- Every <tool> block takes either the form <tool>{"name": str, "args":          
  dict}</tool> or <tool>{"name": str}</tool>, depending on the tool_card.
- The tool results is not produced by you. Do not fabricate any tool output.
- You may use the the given official ViF-CoT-4K reasoning for this sample as a reference for your reasoning and may insert 1-2 additional tool calls from the tool-card block between <answer>, choosing tools that are most relevant to the artifact described.

\end{lstlisting}

\subsection{Student Prompt for SFT, GRPO, and Evaluation}
\label{app:D_student_prompt}

\noindent\textbf{Shared student prompt.} The student receives the same prompt structure during Stage~1, Stage~2, and final evaluation. The prompt includes the full current tool card block; during tool evolution, this block is also updated when tools get updated. After every round’s student model’s output, we run the invoked tools. After the tools return results, we append the tool output and previous messages to the end of this prompt, then send the combined content to the student model to generate the next round of output. The prompt is:
\begin{lstlisting}[style=prompt]
Here are uniformly sampled frames in one video sample:
- frame_idx 0:  <attached image>
- frame_idx 1:  <attached image>
- ... 

Your task is to use the frames and the toolset provided later to reason about whether the video is AI-generated or real. Include <think>, an <answer>, 1-3 code-tool calls from the tool_card block, then toolbox_guide as the last tool call. Stop immediately after toolbox_guide. Reason with these tools and decide; call the tools from the tool cards below that you think can help you further inspect artifacts in the video. (tool calls refer to frames by sampled-frame index starting from 0). 

Below is a toolset, presented as tool cards, that helps you judge whether the given video sample is real or AI-generated.
<tool_cards>
... one compact card per forensic tool ...
</tool_cards>

schema of your output:
<think>...</think>
<answer>Real|Fake</answer>
You could also append with tool call as below:
<tool>tool_name_1</tool>
<tool>tool_name_2</tool>
...
<tool>toolbox_guide</tool>

General rules of your output:
- Only tools in the tool-card block may appear inside <tool> tags.
- Every <tool> block takes either the form <tool>{"name": str, "args":          
  dict}</tool> or <tool>{"name": str}</tool>, depending on the tool_card.
- The tool results is not produced by you. Do not fabricate any tool output.

\end{lstlisting}

\subsection{Tool Evolution Operator Prompts}
\label{app:D_evo}

\noindent\textbf{Mutate prompt.}
\begin{lstlisting}[style=prompt]
[role: evolution operator MUTATE]
This tool currently still has low usage breadth/ low evidential effectiveness. Please read its code carefully and think about how to improve it so that it can cover more edge cases, detect the intended features more reliably, and become a better tool.(e.g. tighten a threshold, restrict to a specific frame subset, swap
aggregation function, add an early exit, gate on cam_motion_classifier
result, OR change the temporal granularity if reasoning supports it).
Optional tool-error section (omitted when no tool error is observed in the
current evolution cycle):
The following tool error messages have been observed so far. You may use
them as concrete debugging evidence:
[error message 0, error message 1, ...]
Return the FULL revised source for the function (not a diff).
\end{lstlisting}

\noindent\textbf{Crossover prompt.}
\begin{lstlisting}[style=prompt]
[role: evolution operator CROSSOVER]
Review the current toolset and their using history, fuse two or more tools that you think are complementary in functionality or case coverage into a new tool. Alternatively, if you think combining the results of two or more tools can produce an effective signal for judging whether a sample is real or fake, you may also create a new tool that combines the results of those tools as a new tool in the toolset.
\end{lstlisting}

\noindent\textbf{Brainstorm prompt.}
\begin{lstlisting}[style=prompt]
[role: evolution operator BRAINSTORM]
Open-ended brainstorming. Based on the current failed samples and their full reasoning chains, create a new tool that you think can capture artifacts in video samples that are still missed by the current toolset.
\end{lstlisting}

\section{Licenses}
\label{app:licenses}
We use ViF-Bench and ViF-CoT-4K~\cite{skyra2026} by following the license of \href{https://huggingface.co/collections/JoeLeelyf/skyra}{here}. We use the GenVideo dataset~\cite{genvideo2024} by following the license of \href{https://github.com/chenhaoxing/DeMamba}{here}. We use the Qwen2.5-VL-7B-Instruct~\cite{qwen25vl2025} by following the license of \href{https://huggingface.co/Qwen/Qwen2.5-VL-7B-Instruct}{here}.

\end{document}